\documentclass[sigconf, nonacm]{acmart}

\AtBeginDocument{%
  \providecommand\BibTeX{{%
    \normalfont B\kern-0.5em{\scshape i\kern-0.25em b}\kern-0.8em\TeX}}}

\copyrightyear{2024}
\acmYear{2024}
\setcopyright{rightsretained}
\acmConference[IVA '24]{ACM International Conference on Intelligent Virtual Agents}{September 16--19, 2024}{GLASGOW, United Kingdom}
\acmBooktitle{ACM International Conference on Intelligent Virtual Agents (IVA '24), September 16--19, 2024, GLASGOW, United Kingdom}\acmDOI{10.1145/3652988.3673934}
\acmISBN{979-8-4007-0625-7/24/09}
\begin{document}



\title{2D or not 2D: How Does the Dimensionality of Gesture Representation Affect 3D Co-Speech Gesture Generation?}

\author{T\'eo Guichoux$^{1,2}$, Laure Soulier$^1$, Nicolas Obin$^2$, Catherine Pelachaud$^{1,3}$}
\affiliation{ \institution{\small
 $^1$ Sorbonne Université, ISIR, F-75005 Paris, France ;\
 $^2$ Sorbonne Université, IRCAM, Stms Lab, F-75003, Paris France ;\
 $^3$ CNRS}
 \country{
    \texttt{
      \{teo.guichoux, laure.soulier, catherine.pelachaud\}@isir.upmc.fr }}}

\renewcommand{\shortauthors}{T.Guichoux, L.Soulier, N.Obin, C.Pelachaud}


\begin{abstract}
    Co-speech gestures are fundamental for communication. The advent of recent deep learning techniques has facilitated the creation of lifelike, synchronous co-speech gestures for Embodied Conversational Agents. "In-the-wild" datasets, aggregating video content from platforms like YouTube via human pose detection technologies, provide a feasible solution by offering 2D skeletal sequences aligned with speech. Concurrent developments in lifting models enable the conversion of these 2D sequences into 3D gesture databases. However, it is important to note that the 3D poses estimated from the 2D extracted poses are, in essence, approximations of the ground-truth, which remains in the 2D domain. This distinction raises questions about the impact of gesture representation dimensionality on the quality of generated motions — a topic that, to our knowledge, remains largely unexplored. Our study examines the effect of using either 2D or 3D joint coordinates as training data on the performance of speech-to-gesture deep generative models. We employ a lifting model for converting generated 2D pose sequences into 3D and assess how gestures created directly in 3D stack up against those initially generated in 2D and then converted to 3D. We perform an objective evaluation using widely used metrics in the gesture generation field as well as a user study to qualitatively evaluate the different approaches.
\end{abstract}




\maketitle

\section{Introduction}

\begin{figure*}
  \centering
  \includegraphics[width=\linewidth]{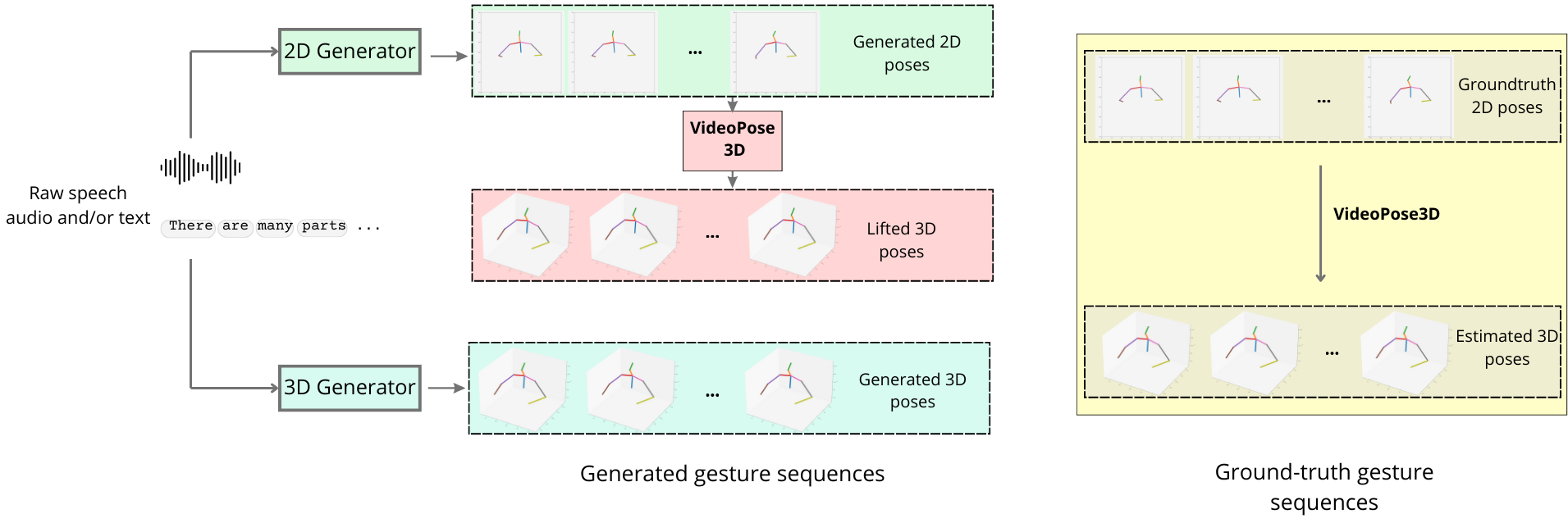}
  \caption{The proposed evaluation pipeline is a combination of a generative model \cite{zhu2023taming, Yoon2020Speech} that generates sequences of 2D body poses and VideoPose3D \cite{pavllo:videopose3d:2019} that lifts the generated 2D poses to 3D. The pseudo-ground-truth 3D gesture sequences originate from the TED Gesture-3D dataset \cite{Yoon2020Speech} and were obtained using VideoPose3D to lift 2D keypoints to 3D. The 2D keypoints were estimated using OpenPose \cite{cao2019Openpose} on TED YouTube videos.}
  \label{fig:pipeline}
\end{figure*}
In human communication, gestures play an integral role by conveying intentions and emphasizing points \cite{mcneill1994hand}.  Recent studies \cite{Alexanderson_2023, yang2023diffusestylegesture, yang2023qpgesture, Ao2023GestureDiffuCLIP, zhu2023taming, alexanderson2020style, Mehta_2023, voss2023aq, deichler2023diffusion, Yoon2020Speech, fares2023zero, yoon2018robot} aim to create similar gestures for Embodied Conversational Agents (ECA) to make interactions with humans more natural and effective. These new methods use learning algorithms and extensive human motion datasets to generate gestures alongside speech. The representation of co-speech gestures, in 2D or 3D, influences how the agent's non-verbal communication is perceived, especially the speaker's communication style \cite{habibie2021learning}.

Multiple learning-based methods for gesture generation focus on generating 2D gestures \cite{fares2023zero,ahuja2020style, ginosar2019learning, qian2021speech}. Data of 2D motion is easily accessible from "in-the-wild" monocular videos, which are videos freely accessible online, therefore allowing the gathering of large-scale collection of motion data, known as "in-the-wild" datasets. However, most of the recent literature considers 3D motion data \cite{Alexanderson_2023, yang2023diffusestylegesture, yang2023qpgesture, Ao2023GestureDiffuCLIP, zhu2023taming, alexanderson2020style, Mehta_2023, voss2023aq, deichler2023diffusion, Yoon2020Speech}, primarily because such data representation contains the depth dimension and is more easily transferable to downstream applications such as 3D virtual agents or social robots \cite{Nyatsanga_2023, yoon2018robot}. But, it is not easy to collect high-quality 3D motion data, as one needs a motion capture setup in a controlled environment, hence limiting the size and diversity of such datasets. To access 3D motion data and still gather large-scale datasets of diverse and spontaneous gestures, multiple works leverage an estimation of the 3D gestures inferred from 2D poses extracted from "in-the-wild" videos \cite{Yoon2020Speech, liu2022learning, voss2023aq}. Nevertheless, to convert extracted 2D keypoints to 3D, one needs a third-party 2D-to-3D lifter, which may be prone to inaccuracies, notably because of the ambiguous nature of 3D pose estimation from 2D keypoints \cite{pavllo:videopose3d:2019}. 

The co-speech gesture generation field has seen a large shift into 3D gesture modeling. However, methods trained on 3D "in-the-wild" datasets still face the bottleneck of the 2D body pose representation. In "in-the-wild" datasets lifted in 3D, the source is still 2D monocular videos. Thus, the ground-truth gestures are in 2D. Deciding whether to convert 2D poses to 3D beforehand or to train a model to generate 2D gestures and then lift them to 3D in post-processing, needs to be addressed. It remains unclear what the impact of the dimensionality of the skeleton representation is on the training of the generative model and the quality of the generated gestures. To the best of our knowledge, this question has never been extensively studied.\\ 

In this work, we compare two training settings to evaluate the influence of data dimensionality on the performance of two speech-to-gesture generative models by either considering 2D or 3D joint coordinates. We use a convolutional neural network \cite{pavllo:videopose3d:2019} to subsequently convert generated 2D gestures to 3D. We chose to evaluate the effect of the pose representation on \textbf{1)} a Denoising Diffusion Probabilistic Model (DDPM) \cite{ho2020diffusion, sohl2015deep, zhu2023taming} and \textbf{2)} a recurrent neural generative model \cite{Yoon2020Speech}. Both approaches have proven their ability to generate natural and diverse gestures aligned with speech. 

There is a one-to-many relationship between 2D keypoints and their 3D counterparts. Given the deterministic nature of the 2D-to-3D lifter, it will consistently map any given 2D pose to the same corresponding 3D pose introducing an inductive bias in the process. We aim to measure whether the gesture distribution resulting from lifting will be less human-like, diverse, and in sync with speech than the distribution of 3D gestures generated via a model trained to directly generate 3D gestures.
More particularly, our contributions are the following:
\begin{itemize}
    \item We propose an evaluation pipeline to investigate the impact of the dimensionality of the pose representation on the performance of two co-speech gesture generative models \cite{zhu2023taming, Yoon2020Speech}. We train both models to generate sequences of body poses represented in 2D coordinates which are then lifted to 3D using VideoPose3D \cite{pavllo:videopose3d:2019}. The pipeline is described in Figure~\ref{fig:pipeline}.
    \item We empirically compare the quality of the gestures generated in 2D lifted to 3D to the gestures directly generated in 3D using evaluation metrics commonly used in co-speech gesture generation tasks \cite{Yoon2020Speech, zhu2023taming, liu2022learning}. 
    \item Additionally, we conducted a user study where participants were asked to choose between gestures generated in 2D then lifted to 3D and gestures directly generated in 3D, providing a direct comparison of perceived quality.
\end{itemize}

The remainder of this paper is organized as follows: first, we present state-of-the-art works focusing on co-speech gesture generation. Secondly, we introduce our methodology and experimental design. Then, we discuss the experimental results of our objective and user studies. We finally end with a conclusion.

\begin{table*}
\caption{Speech-Gesture datasets since 2018. The collection methods are described in the rightmost column. It can be either Motion Capture (MoCap) or pose estimation. Abbreviations: \textit{rot.} \textit{rotations} \textit{coord.} \textit{coordinates}, \textit{n.s} \textit{not specified}. For a more exhaustive list, refer to Nyatsanga et al.\cite{Nyatsanga_2023}.}
\label{tab:dataset}
\small
\begin{tabular}{|l|l|l|l|l|l|}
\hline
Dataset &
  Size &
  \# of speakers &
  Type of motion data &
  Finger motion &
  Collection method
  \\ \hline
TED Gesture, 2018 \cite{yoon2018robot} &
    52.7 h &
    1,295 &
    2D joint coord. &
    &
    OpenPose \cite{cao2019Openpose}  \\

Trinity Speech Gesture I, 2018 \cite{ferstl2018trinity} &
  6h &
  1 &
  3D joint rot. &
  &
  MoCap  \\

SpeechGesture, 2019 \cite{ginosar2019learning}  &
  144 h &
  10 &
  2D joint coord. &
  Yes &
  OpenPose \cite{cao2019Openpose} \\

TalkingWithHands, 2019\cite{lee2019talking} &
    50h &
    50 &
    3D joint rot. &
    Yes &
    Mocap \\
    
TED Gesture 3D, 2020 \cite{Yoon2020Speech}&
  97h &
  n.s. &
  3D joint coord. &
  &
  OpenPose \cite{cao2019Openpose}, 
  VideoPose3D \cite{pavllo:videopose3d:2019} \\

Trinity Speech Gesture II, 2021 \cite{frestl2021trinity2} &
  4h &
  1 &
  3D joint rot. &
  &
  Mocap \\

SpeechGesture 3D , 2021\cite{habibie2021learning} &
  33h &
  6 &
  3D joint coord. &
  Yes &
  OpenPose \cite{cao2019Openpose}, XNect \cite{mehta2019XNect}, \cite{garrido2016reconstruction}\\
  
TED Expressive, 2022 \cite{liu2022learning} &
  100.8h &
  n.s. &
  3D joint coord. &
  Yes &
  OpenPose \cite{cao2019Openpose}, ExPose \cite{ExPose:2020} \\
  
PATS, 2020\cite{ahuja2020no} &
  250h &
  25 &
  2D joint coord. &
  &
  OpenPose \cite{cao2019Openpose}  \\
  
BEAT, 2022  \cite{liu2022beat}&
  76h &
  30 &
  3D joint rot. &
  Yes &
  MoCap  \\
  
ZeroEggs, 2022 \cite{ghorbani2022zeroeggs} &
  2h &
  1 &
  3D joint rot. &
  Yes &
  MoCap \\

BiGe, 2023 \cite{voss2023aq} &
  260h &
  n.s. &
  3D joint coord. &
  Yes &
    OpenPose \cite{cao2019Openpose},
    VideoPose3D\cite{pavllo:videopose3d:2019}  \\   
\hline
\end{tabular}
\end{table*}

\section{Related Work}

\subsection{Learning-based co-speech gesture generation}
The co-speech gesture synthesis field has seen an important shift to deep learning approaches for gesture generation due to their effectiveness in creating natural movements that are well-synchronized with speech, with minimal assumptions \cite{Nyatsanga_2023}. 

Deterministic approaches that directly translate speech to gesture sequences have been proposed. Notable model architectures are multi-layer perceptrons (MLP) \cite{kucherenko2020gesticulator}, convolutional neural networks (CNN) \cite{habibie2021learning}, or transformers \cite{bhattacharya2021speech, windle2023the}. Many considered recurrent architectures \cite{Yoon2020Speech, bhattacharya2021speech, liu2022learning, yoon2018robot} for their ability to capture long-term temporal dependencies to generate body pose sequences. Yoon et al. \cite{yoon2018robot} proposed a sequence-to-sequence model that was trained to generate 2D gesture sequences from the TED Gesture dataset. The generated pose sequences were then lifted to 3D with a learned MLP to be mapped onto a social robot. Later on, Yoon et al. \cite{Yoon2020Speech} introduced Trimodal, a multimodal recurrent neural network (RNN) that translates speech audio and text to 3D gestures conditioned on speaker identity. Their model was trained on the TED Gesture-3D dataset \cite{Yoon2020Speech}.

There is a notable interest in non-deterministic generative models such as Variational Autoencoders (VAEs) \cite{li2021audio} and diffusion models \cite{ho2020diffusion,  song2019generative, Alexanderson_2023, deichler2023diffusion, Ao2023GestureDiffuCLIP, chemburkar2023discrete, tonoli2023gesture, zhao2023diffugesture}  due to their capacity for producing a wide array of gesture types. Specifically, VAEs are designed to encode gestures into a continuous latent space and subsequently decode these latent representations into speech-conditioned movements \cite{li2021audio}.  

Recently, the gesture generation field has particularly focused on Probabilistic Denoising Diffusion Models \cite{ho2020diffusion,  song2019generative, Alexanderson_2023, deichler2023diffusion, Ao2023GestureDiffuCLIP, chemburkar2023discrete, tonoli2023gesture, zhao2023diffugesture} due to their capacity to robustly produce diverse and realistic gestures under multiple conditions, including speech, text, speaker identity, and style. In diffusion-based methods audio-driven gesture synthesis is generally executed through classifier-free guidance \cite{ho2022classifierfree, Alexanderson_2023, zhu2023taming, Ao2023GestureDiffuCLIP}, leveraging both conditional and unconditional generation mechanisms during the sampling process.
Alexanderson et al. \cite{Alexanderson_2023} used Conformers \cite{gulati2020conformer} to generate gestures conditioned on behavior style and speech audio.
Ao et al. \cite{Ao2023GestureDiffuCLIP} leveraged CLIP \cite{clip2021radford} to encode speech text and a style prompt. The authors used a combination of AdaIN \cite{huang2017adain} and classifier-free guidance to generate diverse yet style-conditioned gestures from speech.
Zhu et al. \cite{zhu2023taming} proposed DiffGesture, using a Diffusion Audio-Gesture Transformer to guarantee temporally aligned generation. In their work, raw speech audio is concatenated to gesture frames to condition the diffusion process. DiffGesture is trained on the TED Gesture-3D dataset \cite{Yoon2020Speech} compiling 3D gestures inferred from 2D poses obtained from monocular video. 

\subsection{Representation and collection of the gesture data} 
The quality and diversity of the training data are critical for training co-speech gesture generative models. Additionally, to properly ground gestures to speech audio or text, gathering a large quantity of gesture data paired with these modalities is paramount. 

Early works mostly considered 2D motion data for training \cite{fares2023zero, yoon2018robot, ginosar2019learning, qian2021speech, ahuja2020style}. To collect such datasets, 2D gestures were typically extracted from "in-the-wild" monocular videos using a third-party pose extractor such as OpenPose \cite{cao2019Openpose, yoon2018robot, ahuja2020no, ginosar2019learning}. We report in Table \ref{tab:dataset} a list of existing gesture datasets. This collection process makes it possible to gather a large amount of training data with numerous distinct speakers and ensures the diversity and spontaneity of the gestures. However, leveraging such pre-trained pose estimators induces errors resulting in less expressive motion quality, especially for capturing shoulder and finger movements, and limits the pose representation to be two-dimensional. 

Most of the recent literature \cite{Alexanderson_2023, yang2023diffusestylegesture, yang2023qpgesture, Ao2023GestureDiffuCLIP, liu2022beat, zhao2023diffugesture, kucherenko2023genea} focuses on MoCap datasets \cite{ferstl2018trinity, frestl2021trinity2, liu2022beat, ghorbani2022zeroeggs, lee2019talking}, in which high-quality motion data is captured in a controlled environment, usually with a limited number of speakers, consequently affecting the variety and spontaneity of the gestures.
Multiple works \cite{zhu2023taming, Yoon2020Speech, voss2023aq, liu2022learning, habibie2021learning} opted for increased diversity and volume of data samples while keeping a 3D representation of gestures, and chose to train their models on datasets of 3D gestures collected from "in-the-wild" videos \cite{voss2023aq, Yoon2020Speech, zhu2023taming, habibie2021learning, liu2022learning}. To extract 3D body poses from monocular videos, the data collection process typically leverages a pipeline of pose extraction \cite{cao2019Openpose} and 2D-to-3D lifting \cite{pavllo:videopose3d:2019, ExPose:2020, mehta2019XNect}. For instance, the dataset TED Gesture-3D introduced by Yoon et al. \cite{Yoon2020Speech} leverages VideoPose3D \cite{pavllo:videopose3d:2019} to convert 2D body keypoints extracted by OpenPose \cite{cao2019Openpose} to 3D. This dataset is an extension of the previous TED Gesture dataset \cite{yoon2018robot} where the poses are represented in 2D. \footnote{To avoid confusion we refer to the 3D version of the database used by \cite{Yoon2020Speech} as \textit{TED Gesture-3D}.}.

There has been an important shift in the field towards 3D gestures generation \cite{Alexanderson_2023, yang2023diffusestylegesture, yang2023qpgesture, Ao2023GestureDiffuCLIP, liu2022beat, zhao2023diffugesture, zhu2023taming, Yoon2020Speech, voss2023aq, liu2022learning, habibie2021learning}. However, the impact of the pose representation - either 2D or 3D - on the quality of synthesis gestures remains largely unexplored. Kucherenko et al. studied the impact of motion representation on the performances of data-driven co-speech gesture generative models, but their work only focused on the study of a gesture representation learned in a latent space.
\\

In this work, we study how training an audio-driven generative model to synthesize 2D motion data and then post-processing the generated sequences using a 3D lifter impacts the overall quality of the synthesized gestures. We choose to use DiffGesture \cite{zhu2023taming} and Trimodal \cite{Yoon2020Speech} as baselines for our study as both were initially trained on the TED Gesture-3D dataset and obtained good performances, both qualitatively and objectively. Both models belong to widely used classes of generative models, DDPMs \cite{zhao2023diffugesture, Alexanderson_2023, tonoli2023gesture, Ao2023GestureDiffuCLIP, chemburkar2023discrete} and recurrent encoder-decoders \cite{liu2022learning, voss2023aq, yoon2018robot}. For the 2D-to-3D lifting model, we employ VideoPose3D \cite{pavllo:videopose3d:2019}. We use the TED Gesture-3D dataset \cite{Yoon2020Speech} for our evaluation, which is a dataset of 3D gestures extracted from YouTube videos, hence covering a large array of speakers with different gesturing styles. 


\section{Methodology}


\subsection{TED Gesture-3D dataset}

 TED Gesture-3D \cite{Yoon2020Speech} is a dataset including pose sequences extracted from in-the-wild videos of TED talkers with the corresponding speaker identity, speech audio, and speech transcription. TED Gesture-3D includes 3D body poses estimated via a combination of a 2D pose extractor from monocular videos \cite{cao2019Openpose} and VideoPose3D \cite{pavllo:videopose3d:2019}.
The size of the dataset is 97h where the poses are sampled at 15 frames per second with a stride of 10 with a total of 252,109 sequences of 34 frames. TED Gesture-3D is divided into training, validation and test sets which respectively represent 80\%, 10\% and 10\% of the total dataset.
Body poses are represented as vectors in $\mathcal{R}^{N\times J \times 3}$ where $N$ is the sequence length and $J$ is the number of body joints. Instead of considering raw joint coordinates for body pose representation, we follow the approach proposed by Yoon et al. \cite{Yoon2020Speech} where a body pose is represented as nine directional vectors where each direction represents a bone. The vectors are normalized to the unit length and centered on the root joint. This pose representation is invariant to bone length and less affected by root rotations therefore favoring the training.
Regarding 2D pose sequences, they are vectors of 3D poses from which the depth axis has been removed. 

\subsection{Pipeline}
\label{pipeline}
To evaluate the inductive bias caused by the dimensionality of the gesture representation (2D or 3D) and the 2D-to-3D conversion, we trained the co-speech gesture generators of \cite{Yoon2020Speech} and \cite{zhu2023taming} on both 2D and 3D settings. We employed a 3D lifter for post-processing the 2D generated sequences to be able to compare them to the 3D generated sequences. The complete pipeline is described in Figure~\ref{fig:pipeline}. 

\subsubsection{Gesture generators}\ We now present the two gesture generators used as references in this study: DiffGesture \cite{zhu2023taming} and Trimodal \cite{Yoon2020Speech}. \\

\noindent \textbf{1) DiffGesture} is defined as a DDPM that generates sequences of poses out of noise, conditioned on raw speech audio. DDPMs rely on two Markov chains: the forward process that gradually adds noise to the data and the backward process that converts noise to data. In DiffGesture, the backward process is modeled as a deep neural network that synthesizes gestures conditioned on speech. Raw audio is encoded using a convolutional neural network and then concatenated to the noisy pose sequence along the features axis.
To synthesize diverse and speech-accurate gestures, DiffGesture uses classifier-free guidance \cite{ho2022classifierfree}. This approach involves jointly training a conditioned and an unconditioned DDPM, allowing for a trade-off between the quality and diversity of the generated poses at inference time. \\
\noindent \textbf{2) Trimodal} is an encoder-decoder model trained in an adversarial scheme, that translates speech audio and text into 3D gestures, conditioned on speaker identity. It employs three distinct neural network encoders to process the three input modalities: audio, text, and speaker identity. The gesture generation utilizes a bi-directional gated recurrent unit (GRU) \cite{Bahdanau2014NeuralMT} to maintain temporal consistency. Audio and text inputs are handled by separate convolutional networks, while speaker identity is encoded into a style vector using a VAE. This VAE constructs a latent `style' embedding space that captures the unique characteristics of each speaker. The style feature vector derived from this space is consistently applied across all synthesis time steps, ensuring coherent gesture representation throughout the sequence.
\\

DiffGesture and Trimodal were first designed to generate 3D gestures. We adapted these architectures to account for 2D body pose sequences by changing the input and output dimensions of the denoising network and recurrent decoder network respectively. Specifically, we removed the depth axis of the body pose coordinates thus considering poses in $\mathcal{R}^{J\times2}$. We refer to these versions as \textit{DiffGesture 2D} and \textit{Trimodal 2D} and the original versions are referred to as \textit{DiffGesture 3D} and \textit{Trimodal 3D}.
We trained DiffGesture and Trimodal in two different settings: 2D motion generation and 3D motion generation, as described in Figure~\ref{fig:pipeline}. For all experiments, we follow the original implementation proposed by Zhu et al. \cite{zhu2023taming} and Yoon et al. \cite{Yoon2020Speech}. In our objective study, we obtained similar results as Zhu et al. and Yoon et al. when retraining \textit{DiffGesture 3D} and \textit{Trimodal 3D} demonstrating the validity of our evaluation protocol (see Table \ref{tab:experimental-results}). 

\subsubsection{2D-3D Lifter}\ We employed a 2D-3D lifter defined by a temporal convolutional network (TCN). Specifically, we used VideoPose3D \cite{pavllo:videopose3d:2019} to lift 2D pose sequences to 3D as illustrated in Figure~\ref{fig:pipeline}. The lifting process is defined as a mapping problem, in which the TCN employs 1-D convolutions along the temporal axis to transform 2D full body poses into a temporally consistent sequence of 3D body poses. VideoPose3D utilizes dilated temporal convolutions to capture long-term information.

We retrained VideoPose3D \cite{pavllo:videopose3d:2019} on the TED Gesture-3D dataset to be able to input body poses in $\mathcal{R}^{2\times 9}$ i.e when only the upper part of the body is considered. We obtained a slightly better mean per joint positional error (MPJPE) when the sequences were up-scaled to 273 frames to exceed the receptive field of VideoPose3D instead of the original 34 frames. The final MPJPE of VideoPose3D was $\boldsymbol{13.4}$ on the test set of TED Gesture-3D. We kept the model architecture and training hyper-parameters consistent with the original implementation. 
\\

Supplementary videos used in the subjective evaluation study are provided at the following website: 
\\
\noindent \href{https://sites.google.com/view/iva-2d-or-not-2d}{{\fontfamily{qcr}\selectfont https://sites.google.com/view/iva-2d-or-not-2d}}.

\begin{figure*}
  \centering
  \includegraphics[width=\linewidth]{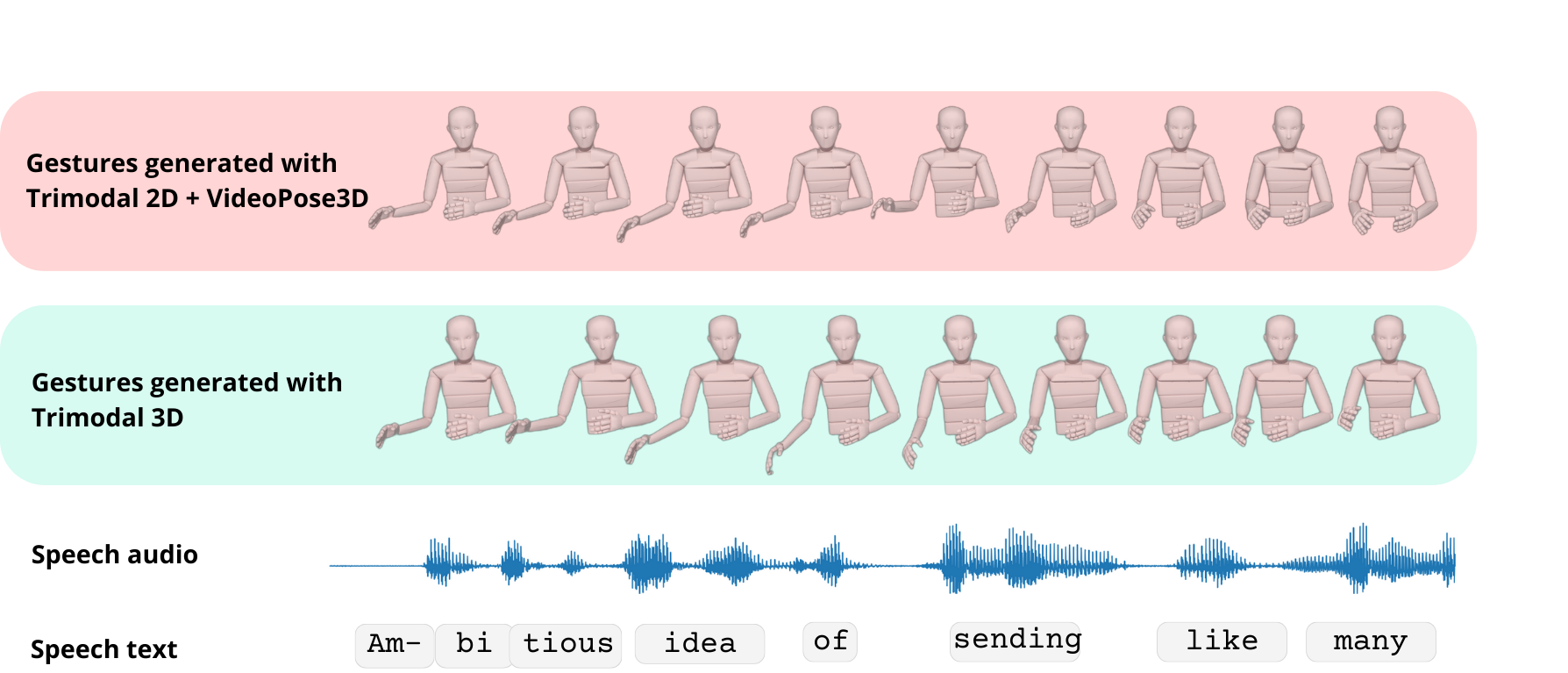}
  \caption{Keyframes of an animation generated with \textit{Trimodal 2D + VP3D} (up) and \textit{Trimodal 3D} (down).}
  \label{fig:keyframes}
\end{figure*}

\subsection{Comparative setting}

\label{sec:experiments}
To study the impact of the dimensionality of the motion data on the performance of gesture generative models, we considered two experimental settings. 

\subsubsection{Evaluation in the 3D gesture space}\ To evaluate the impact of training on 2D motions on the quality of 3D gesture sequences, we define \textit{DiffGesture 2D + VP3D} and \textit{Trimodal 2D + VP3D} as DiffGesture and Trimodal trained on 2D motion data whose outputs are then lifted to 3D using VideoPose3D and we compare them to the original models, \textit{DiffGesture 3D} and \textit{Trimodal 3D} \cite{zhu2023taming, Yoon2020Speech}.

\subsubsection{Evaluation in the 2D gesture space}\ To further explore the impact of motion dimensionality on the generated gesture, we also compare \textit{DiffGesture 2D} and \textit{Trimodal 2D} to \textit{DiffGesture 3D} and \textit{Trimodal 3D} but where the 3D generated motion is narrowed to 2D by removing the depth axis, we refer to these models as \textit{DiffGesture 3D$\rightarrow$2D} and \textit{Trimodal 3D$\rightarrow$2D}.

\subsection{Objective Evaluation metrics}
\label{sec:metrics}

We numerically evaluate our models with three commonly used metrics in the co-speech gesture generation field. 

\subsubsection{The Fréchet Gesture Distance} The Fréchet Gesture Distance (FGD) defined by Yoon et al. \cite{Yoon2020Speech} is an adaptation of the Fréchet Inception Distance (FID) \cite{gans2017Heusel}. The FGD computes the 2-Wasserstein distance between two distributions leveraging latent features extracted with a pose encoder. Similar distributions result in a low FGD value. The FGD is defined as follows:
\begin{equation}
    FGD(X,\hat{X}) = \lVert \mu_r - \mu_g \rVert + Tr(\Sigma_r + \Sigma_g - 2(\Sigma_r \Sigma_g)^2 )
\end{equation} 
Where: $X$, $\hat{X}$ are the real and generated distributions respectively; $\mu_r$, $\mu_g$, $\Sigma_r$, $\Sigma_g$ are the mean and covariance of the latent distributions extracted from the real and generated distributions.

\subsubsection{The Beat Consistency Score} The Beat Consistency Score (BC) measures the temporal consistency between kinematic and audio beats of a paired audio-motion sequence. This measure, first introduced for evaluating the synchrony of dance with music \cite{dance2021Li}, has been adapted to speech and gestures \cite{learn2021Li}.
First, kinematic beats are extracted from a pose sequence by selecting the time steps of the sequence where the average angle velocity is higher than a certain threshold. Angle velocity is computed using the variation in angle between two successive frames. Intuitively, BC measures the average distance between the time steps corresponding to an audio beat and the closest time steps corresponding to a kinematic beat. The audio beats are extracted using a pre-trained detection model and the BC score is computed as follows:
\begin{equation}
\label{eq:bc}
     BC = \frac{1}{n}\sum_{i=0}^n exp \Big ( - \frac{min_{\forall t_{j}^y \in \mathcal{B}^y} \lVert t_{j}^y - t_{i}^x \rVert ^2}{2\sigma^2} \Big )
\end{equation}
Where: $t_{i}^x$ is the $i-th$ audio beats, $B_y = {t_{i}^y }$ is the set of the kinematic beats of the $i-th$ sequence, and $\sigma$ is a parameter to normalize sequences, set to 0.1 empirically as in \cite{zhu2023taming}. 

\subsubsection{The Diversity measure} The Diversity measure also leverages the latent features extracted with a pose encoder \cite{hsin2019diversity}. Diversity is computed by randomly selecting two sets of $N$ features from the generated distribution and calculating the distance between the mean of both sets in the feature space.  Typically, if a model generates similar gestures all gestures will be close to the average gesture sequence, resulting in a small distance between the two sets, as formalized below:
\begin{equation}
\label{eq:diversity}
    Div(X) = \lVert \mu_{A} - \mu_{B} \rVert_2 
\end{equation}
Where: $X$ is a distribution of gestures, $A$ and $B$ are sets of gestures randomly sampled from $X$, and $\mu_A$ and $\mu_B$ are the mean of the gesture features in both sets.
\\

The FGD and diversity metrics are calculated using an auto-encoder designed to encode 3D gestures into latent space. Yoon et al. \cite{Yoon2020Speech} developed this auto-encoder using the human3.6m dataset \cite{h36m_pami}. Similarly, we adapted this model to encode 2D gestures by eliminating the depth axis and training it on the same dataset.

\section{User Study}

Properly evaluating generative models is a challenging task, especially in the co-speech gesture generation field partly because of the subjective nature of human communication \cite{Yoon2020Speech, Nyatsanga_2023, Alexanderson_2023}. In this section, we present our user study protocol to qualitatively compare the gestures directly generated in 3D to those generated in 2D and subsequently lifted to 3D.

\subsection{Protocol}
We created videos showing the animation of the upper body of an articulated humanoid. Each video features two stimuli for pairwise comparison as it has been shown to reduce the cognitive load of the users \cite{Wolfert_2022, Yoon2020Speech, clark2018whyrate}. The first animation is displayed on the left side of the screen with the second animation masked. The second animation is shown while the first is masked. In each video, both animations used the same model (either DiffGesture or Trimodal), one with direct 3D gestures and the other with 2D gestures converted to 3D. We qualitatively evaluated the impact of the 2D-to-3D lifting, by including baseline videos  (referred to as \textit{Human GT}). These videos paired 3D pseudo-ground-truth gestures to those created by converting the 2D versions of these gestures to 3D using the retrained version of VideoPose3D. The pseudo-ground-truth gestures originate from the test set of the TED Gesture-3D dataset.
After viewing each video, participants were asked to answer three questions:
\begin{itemize}
    \item "\textit{Select the video in which the articulated figure is more \textbf{human-like}}."
    \item "\textit{Select the video in which the articulated figure looks more \textbf{alive}}."
    \item "\textit{Select the video in which the articulated figure looks more \textbf{in sync with the speech}}."
\end{itemize}

We selected the terms "human-like" and "alive" to evaluate two distinct dimensions: the anthropomorphism and animacy of the agent, following the semantics of the Godspeed Questionnaire \cite{Bartneck2009}.
For each question, there were four possible answers: "\textit{Clearly left}", "\textit{Fairly left}", "\textit{Fairly right}", and "\textit{Clearly right}". The options "Clearly" and "Fairly" correspond to the degree of confidence the participants had in their choice.
Each response is assigned an integer value: +2 for a clear preference for gestures directly generated in 3D, +1 for a slight preference, -1 for a slight preference for lifted 3D gestures, and -2 for a clear preference for lifted 3D gestures.

In our subjective study, we created 14 stimuli for each condition: \textit{Human GT}, \textit{DiffGesture}, and \textit{Trimodal}, resulting in 42 pairwise comparisons of the 3D model and its 2D counterpart. To reduce the length of the questionnaires and keep the participants focused during the study, we conducted two evaluation sessions featuring 21 stimuli. Hence each participant saw 7 stimuli for each condition. To prevent ordering bias, we created four distinct questionnaires for each session where the order of appearance of each stimuli has been randomized. In addition to the main stimuli, we included a first example video to familiarize the users with the task and check for technical issues such as no audio or video.
Our questionnaire also featured two attention checks. In the first attention check, the stimuli were replaced by a black-screen video and users were asked to select a specific option three times instead of the questions about human likeness, aliveness, and speech synchrony. In the second attention check, we replaced the original stimuli with a modified version where the audio indicated the users to choose the rightmost option for each question.

\subsection{Gestures rendering}
For visualization, we used the 3D character from the user evaluation in the Genea Challenge 2020 \cite{kucherenko2020genea}. The gestures were rendered with Blender.
To create the animations, we selected 14 speech samples from the test set of the TED Gesture-3D dataset. Each sample lasts around 10 seconds, which is twice as long as the samples used for the subjective evaluation in Yoon et al. \cite{Yoon2020Speech}, Zhu et al. \cite{zhu2023taming} did not report the duration of their stimuli. We selected the samples based on the audio quality and for each sample, we generated four 3D co-speech gesture sequences with \textit{DiffGesture 3D}, \textit{DiffGesture 2D + VP3D}, \textit{Trimodal 3D}, and \textit{Trimodal 2D + VP3D}. As a baseline, we also included gesture sequences directly from the TED Gesture-3D dataset and those were reduced to 2D and then converted back to 3D. Hence, for each audio sample, we obtained 6 sequences for the gestures generated in 3D and their 2D-to-3D counterpart. We created videos that pair the gestures generated in 3D and those converted from 2D, resulting in 42 animation pairs. The order of animations (either direct 3D gestures first or lifted gestures first) was randomized to prevent ordering bias. 

\subsection{Participants}
The participants in the evaluation were recruited online on the Prolific platform \cite{Douglas2023-vo, Jonell_2020}. Among the 67 participants who took the test, 7 failed the attention checks. The users were 36.7+/-11.8 years old and there were 37 females and 30 males and the median completion time was 17 minutes. As we performed two sessions with 7 stimuli for each condition, we obtained 30 valid responses for each stimulus.
The participants were paid 3\pounds \ if they passed the attention checks.

\section{Experimental results}
\subsection{Objective evaluation}

\begin{table}
\caption{Objective results of the experiments on the TED Gesture-3D dataset \cite{Yoon2020Speech}. These results correspond to the experiments (1) and (2) (see section \ref{sec:experiments}).
Up arrows indicate that a higher result is better whereas down arrows indicate that a lower result is better. * means that the results are reported from \cite{zhu2023taming}.}
\label{tab:experimental-results}
\begin{tabular}{lccc}
\hline
\multicolumn{1}{c}{} &
  \multicolumn{3}{c}{TED Gesture }  \\ \cline{2-4} 
Methods &
  FGD $\downarrow$ & 
  BC $\uparrow$ & 
  Diversity $\uparrow$ 
  \\
\hline
\hline
Evaluation on the 3D gesture space & & \\
\hline

\textbf{Ground Truth 3D} &
    0 & 
    0.702 & 
    102.339  \\
DiffGesture 3D \cite{zhu2023taming} &
  1.947 & 
  0.678 & 
  101.436  \\
DiffGesture 2D + VP3D &
  3.121 &
  0.551 &
  100.822 \\

Trimodal 3D \cite{Yoon2020Speech} &
  3.964 & 
  0.733 & 
  95.253  \\

Trimodal 2D + VP3D &
  6.374 &
  0.610 &
  93.017 \\

\hline
Evaluation on the 2D gesture space & & \\
\hline

\textbf{Ground Truth 2D} &
  0 &
  0.689 &
  112.76 \\

DiffGesture (3D$\rightarrow$2D) &
  2.452 & 
  0.661 &
  109.978  \\
  
DiffGesture 2D &
  2.971 & 
  0.644 & 
  111.869 \\

Trimodal (3D$\rightarrow$2D) &
  5.295 & 
  0.724 & 
  104.072 \\

Trimodal 2D &
  6.227 &
  0.706 & 
  102.100 \\

\hline
Reported results from \cite{zhu2023taming}  & &\\
\hline

\textbf{Ground Truth 3D} &
    0 &
    0.698 &
    108.525  \\
DiffGesture * \cite{zhu2023taming} &
  1.506 &
  0.699 &
  106.722 \\
Attention Seq2Seq* \cite{yoon2018robot} &
  18.154 &
  0.196 &
  82.776  \\
Speech2Gesture* \cite{ginosar2019learning}&
  19.254 &
  0.668 &
  93.802  \\
Joint Embedding* \cite{ahuja2019language}&
  22.083 &
  0.200 &
  90.138  \\
Trimodal* \cite{Yoon2020Speech}&
  3.729 &
  0.667 &
  101.247 \\
HA2G* \cite{liu2022learning}&
  3.072 &
  0.672 &
  104.322 \\

\hline
\end{tabular}
\end{table}

The results of our objective experiments are reported in Table \ref{tab:experimental-results}. In the table's upper section, we present outcomes from our experiments evaluating gestures in 3D. The middle section details the results from our experiments assessing gestures in 2D. The results from Zhu et al. and Yoon et al. \cite{zhu2023taming, Yoon2020Speech} are reported in the table's lower section. It is important to note that we retrained the motion encoder used to compute the FGD and diversity score. The reported results from Zhu et al. and Yoon et al. were obtained using their encoder.
\\

\subsubsection{Evaluation of lifted generated gestures} \ When comparing the results of \textit{DiffGesture 2D + VP3D} and \textit{Trimodal 2D + VP3D }to those of \textit{DiffGesture 3D} and \textit{Trimodal 3D}, we can notice that the models trained on 2D gestures perform worse than the original 3D models in terms of FGD, and BC. There is also a slight drop in diversity for Trimodal. We assume that the one-to-many relationship between 2D and 3D keypoints is mostly responsible for the performance drop of \textit{DiffGesture 2D + VP3D} and \textit{Trimodal 2D + VP3D} for the FGD. As VideoPose3D is deterministic, to one 2D pose it will systematically predict the same 3D pose although there exists multiple possibilities. Hence, the distribution resulting from lifting 2D sequences is tighter than the distribution directly generated in 3D, explaining the high FGD of the gestures generated in 2D lifted to 3D.
There is a drop in BC between the 3D models and their 2D counterparts. It can be that post-processing 2D gestures using VideoPose3D tends to over-smooth the resulting 3D gestures therefore reducing the number of kinematic beats.
While the overall quality diminishes when generating gestures in 2D and then lifting them to 3D, \textit{DiffGesture 2D + VP3D} and \textit{Trimodal 2D + VP3D} remain competitive compared to the other baselines reported in the table lower section.
\\

\subsubsection{Evaluation of the quality of gestures generated in 2D} \ When evaluating in the 2D motion space, both \textit{DiffGesture 3D$\rightarrow$2D} and \textit{Trimodal 3D$\rightarrow$2D} perform better than \textit{DiffGesture 2D} and \textit{Trimodal 2D} respectively in terms of FGD. Hence, training the generative models to generate 3D motion sequences seems to behave better than training the model on 2D motion data. This outcome was anticipated since the representation of poses in 3D is richer than the 2D version. 
The BC and diversity scores do not seem to be influenced by the dimensionality of the gestures used to train the generative models.

\subsection{Subjective evaluation}
\begin{figure}
  \centering
  \includegraphics[width=\linewidth]{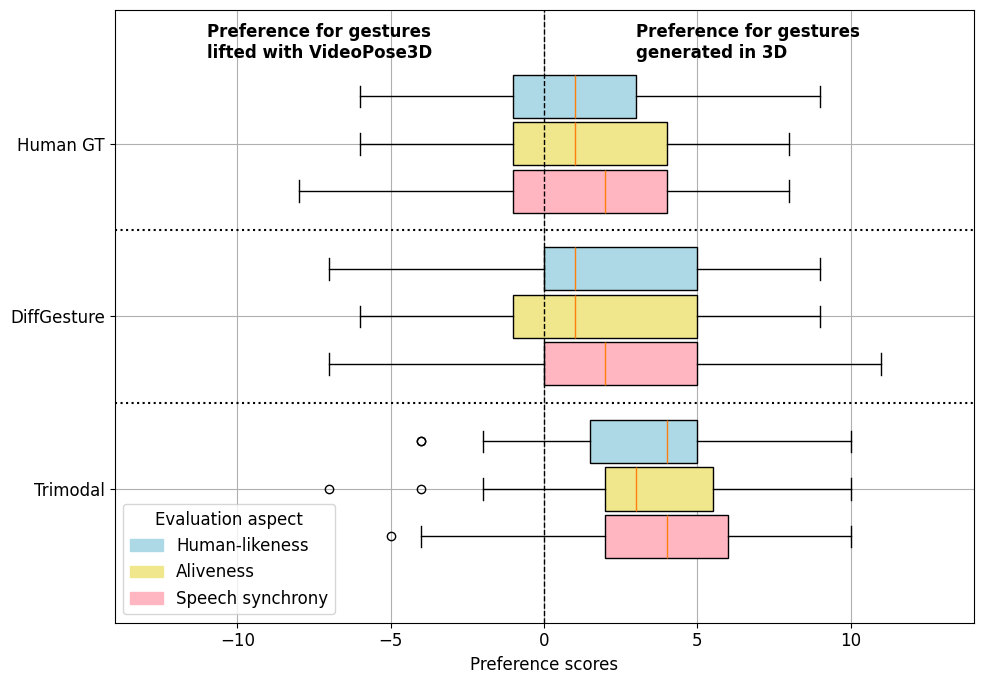}
  \caption{Results of our evaluation study. A positive score means that gestures generated directly in 3D are preferred over 2D gestures lifted to 3D. Reciprocally, a negative score means that 2D gestures lifted to 3D are preferred over direct 3D gestures. A score close to 0 means that the preference is unclear.}
  \label{fig:human_results}
\end{figure}

\begin{figure}[t]
  \centering
  \includegraphics[width=\linewidth]{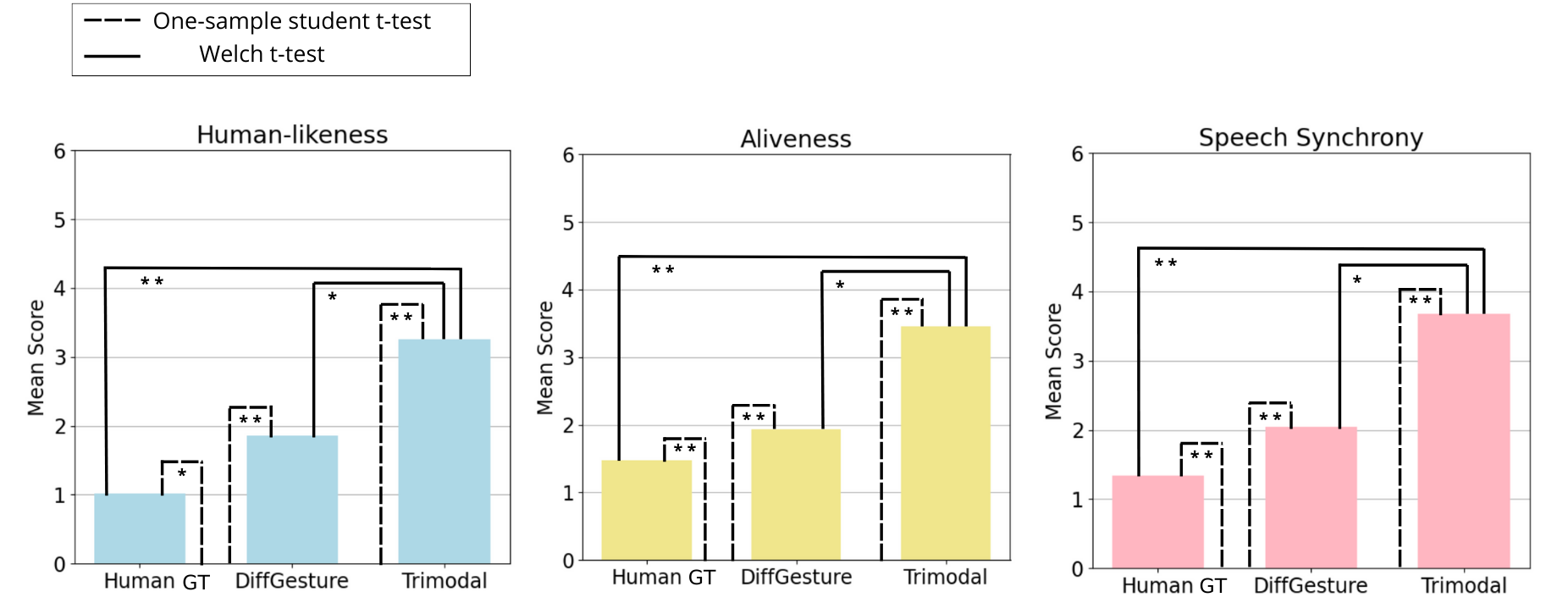}
  \caption{Statistical comparison of mean scores for each model and each aspect (Human-likeness, Aliveness, and Speech synchrony). The lines represent a significant superiority between the two values. Dotted lines correspond to Student t-tests and plain lines to Welch t-tests. \textbf{*} means p-value < 0.05 while \textbf{**} means p-value < 0.01}
  \label{fig:comparison_stat}
\end{figure}

\begin{table}[t]
\centering
\caption{Confidence Intervals of the Mean Scores}
\label{tab:interval}
\begin{tabular}{@{}lccc@{}}
\toprule
Approach & Human-likeness & Aliveness & Speech Synchrony \\ 
\midrule
\textbf{Human GT} & [-0.22, 2.26] & [0.29, 2.66] & [0.08, 2.56] \\
\textbf{DiffGesture} & [0.59, 3.14] & [0.68, 3.18] & [0.83, 3.27] \\
\textbf{Trimodal} & [2.27, 4.23] & [2.46, 4.46] & [2.61, 4.75] \\
\bottomrule
\end{tabular}
\end{table}

The results of our user study are presented in Figure \ref{fig:human_results}. The figure highlights pairwise preference between gestures generated directly in 3D or gestures generated in 2D lifted to 3D using VideoPose3D.
We conducted a statistical analysis to determine the significance of our user study results. We used Student t-tests to evaluate the three different techniques (Human GT, DiffGesture, Trimodal) for generating gestures in 3D. The Student t-test checked if the average scores for human likeness, aliveness, and speech synchrony were significantly different from zero. A score significantly greater than zero indicates that gestures created directly in 3D are better perceived than those initially generated in 2D and subsequently converted to 3D using VideoPose3D. We also conducted Welch t-tests to determine the significance level of the model-wise mean score comparisons. The results of the Student t-tests and Welch tests are depicted in Figure \ref{fig:comparison_stat}.
We calculated the confidence intervals of the mean scores for Human GT, DiffGesture, and Trimodal for the human-likeness, aliveness and speech synchrony. These intervals are depicted in Table \ref{tab:interval}. \\

First, for the DiffGesture and Trimodal techniques, all scores are significantly higher than zero (Figure \ref{fig:comparison_stat}), with a p-value less than 0.01. This suggests that gestures created directly in 3D by these methods are more effective compared to those initially created in 2D and then converted to 3D for all three aspects. In Table \ref{tab:interval}, the confidence intervals of the score of Trimodal show that \textit{Trimodal 3D} is preferred over \textit{Trimodal 2D + VP3D} whereas the preference for \textit{DiffGesture 3D} over \textit{DiffGesture 2D + VP3D} is slighter. 
It is important to note that the confidence intervals of the scores overlap between DiffGesture and Trimodal, but the human-likeness, aliveness, and speech-synchrony means of Trimodal are significantly greater than those of DiffGesture with a p-value of 0.028, 0.016 and 0.011 respectively.
A comparative example between gestures generated with \textit{Trimodal 3D} and \textit{Trimodal 2D + VP3D} is depicted in Figure \ref{fig:keyframes}.

For the Human baseline, converting 2D gestures to 3D demonstrates a minimal yet statistically significant impact on the perception of their human-likeness animation, with a score above zero (p-value of 0.020) and a confidence interval of the mean score close to zero. This suggests that while VideoPose3D influences the perceived human likeness, the effect is subtle. In contrast, the conversion process significantly affects gesture quality in terms of aliveness and speech synchrony, as evidenced by scores significantly higher than zero (p-values of 0.001 and 0.004, respectively). This indicates a notable degradation in these aspects due to the use of VideoPose3D for lifting 2D gestures to 3D.

 From these results, we can conclude that training a model to generate 2D gestures and then converting these gestures to 3D deteriorates the overall animation quality in terms of human likeness, aliveness, and speech synchrony. The 2D-to-3D conversion of gestures has a small yet significant impact on the perception of human-likeness. Hence, the drop in human-likeness quality for gestures generated in 2D and then lifted to 3D may come from the training of the generative model itself since the 2D gesture representation may not allow the generation of highly human-like gestures once converted to 3D. 
 

\section{Conclusion}
In this study, we explored how training two co-speech gesture generators with 2D data affects the overall performance of the models and the perceived quality of the synthesized gestures. We introduced a pipeline that pairs a gesture generator — either DiffGesture \cite{zhu2023taming} or Trimodal \cite{Yoon2020Speech} — with a 2D-to-3D lifting model \cite{pavllo:videopose3d:2019}. Our objective results reveal that using this pipeline negatively impacts overall performance. 3D gestures lifted from 2D generated gestures are less similar to the target 3D gesture distribution in comparison to gestures generated directly in 3D and lifting 2D gestures reduces the BC score.
To further confirm these results, we performed a large-scale user study involving 60 participants. The goal of this human evaluation was to assess the impact of 2D motion representation on the perceived human likeness, aliveness, and speech synchrony of the generated gestures. Our findings show that direct 3D gestures are preferred over gestures lifted using VideoPose3D for both Trimodal \cite{Yoon2020Speech} and DiffGesture \cite{zhu2023taming}. Our findings also indicate that converting 2D gestures to 3D slightly reduces their perceived human likeness, aliveness, and speech synchrony. These results suggest that generating 2D body poses and then lifting them in 3D produces gesture animations that are less human-like, alive, and in sync with speech than those created directly in 3D. Further, they show that using 3D representations for co-speech gesture generation enhances their quality and relation to speech. 



\section{Limitations and future work}
It is worth noting that our study is not without limitations.
In the TED Gesture-3D dataset, the 3D gestures are lifted from 2D body poses. Our evaluation is therefore biased as we do not have access to the real 3D ground truth data. For future research, we plan to conduct a similar analysis using Mocap data to have access to the real 3D ground-truth gestures. 

In the near future, we will test the generalization of our approach on similar "in-the-wild" datasets such as PATS, SpeechGesture and TED Expressive \cite{ahuja2020no, habibie2021learning, ginosar2019learning, liu2022learning}. Additionally, we will consider finger motions which convey a lot of information in human communication \cite{mcneill1994hand}. 
Transforming 2D finger motion into 3D is a complex task, and our focus will be on exploring the best data representation for accurately generating such fine-grained gestures.





\bibliographystyle{ACM-Reference-Format}
\bibliography{references}


\begin{thebibliography}{58}


\ifx \showCODEN    \undefined \def \showCODEN     #1{\unskip}     \fi
\ifx \showDOI      \undefined \def \showDOI       #1{#1}\fi
\ifx \showISBNx    \undefined \def \showISBNx     #1{\unskip}     \fi
\ifx \showISBNxiii \undefined \def \showISBNxiii  #1{\unskip}     \fi
\ifx \showISSN     \undefined \def \showISSN      #1{\unskip}     \fi
\ifx \showLCCN     \undefined \def \showLCCN      #1{\unskip}     \fi
\ifx \shownote     \undefined \def \shownote      #1{#1}          \fi
\ifx \showarticletitle \undefined \def \showarticletitle #1{#1}   \fi
\ifx \showURL      \undefined \def \showURL       {\relax}        \fi
\providecommand\bibfield[2]{#2}
\providecommand\bibinfo[2]{#2}
\providecommand\natexlab[1]{#1}
\providecommand\showeprint[2][]{arXiv:#2}

\bibitem[Ahuja et~al\mbox{.}(2020a)]%
        {ahuja2020no}
\bibfield{author}{\bibinfo{person}{Chaitanya Ahuja}, \bibinfo{person}{Dong~Won Lee}, \bibinfo{person}{Ryo Ishii}, {and} \bibinfo{person}{Louis-Philippe Morency}.} \bibinfo{year}{2020}\natexlab{a}.
\newblock \showarticletitle{No Gestures Left Behind: Learning Relationships between Spoken Language and Freeform Gestures}. In \bibinfo{booktitle}{\emph{Proceedings of the 2020 Conference on Empirical Methods in Natural Language Processing: Findings}}. \bibinfo{pages}{1884--1895}.
\newblock


\bibitem[Ahuja et~al\mbox{.}(2020b)]%
        {ahuja2020style}
\bibfield{author}{\bibinfo{person}{Chaitanya Ahuja}, \bibinfo{person}{Dong~Won Lee}, \bibinfo{person}{Yukiko~I. Nakano}, {and} \bibinfo{person}{Louis-Philippe Morency}.} \bibinfo{year}{2020}\natexlab{b}.
\newblock \showarticletitle{Style Transfer for Co-Speech Gesture Animation: A Multi-Speaker Conditional-Mixture Approach}.
\newblock
\urldef\tempurl%
\url{https://arxiv.org/abs/2007.12553}
\showURL{%
\tempurl}


\bibitem[Ahuja and Morency(2019)]%
        {ahuja2019language}
\bibfield{author}{\bibinfo{person}{Chaitanya Ahuja} {and} \bibinfo{person}{Louis{-}Philippe Morency}.} \bibinfo{year}{2019}\natexlab{}.
\newblock \showarticletitle{Language2Pose: Natural Language Grounded Pose Forecasting}.
\newblock \bibinfo{journal}{\emph{CoRR}}  \bibinfo{volume}{abs/1907.01108} (\bibinfo{year}{2019}).
\newblock
\showeprint[arXiv]{1907.01108}
\urldef\tempurl%
\url{http://arxiv.org/abs/1907.01108}
\showURL{%
\tempurl}


\bibitem[Alexanderson et~al\mbox{.}(2020)]%
        {alexanderson2020style}
\bibfield{author}{\bibinfo{person}{Simon Alexanderson}, \bibinfo{person}{Gustav~Eje Henter}, \bibinfo{person}{Taras Kucherenko}, {and} \bibinfo{person}{Jonas Beskow}.} \bibinfo{year}{2020}\natexlab{}.
\newblock \showarticletitle{Style-Controllable Speech-Driven Gesture Synthesis Using Normalising Flows}.
\newblock \bibinfo{journal}{\emph{Computer Graphics Forum}} \bibinfo{volume}{39}, \bibinfo{number}{2} (\bibinfo{year}{2020}), \bibinfo{pages}{487--496}.
\newblock
\urldef\tempurl%
\url{https://doi.org/10.1111/cgf.13946}
\showDOI{\tempurl}
\showeprint{https://onlinelibrary.wiley.com/doi/pdf/10.1111/cgf.13946}


\bibitem[Alexanderson et~al\mbox{.}(2023)]%
        {Alexanderson_2023}
\bibfield{author}{\bibinfo{person}{Simon Alexanderson}, \bibinfo{person}{Rajmund Nagy}, \bibinfo{person}{Jonas Beskow}, {and} \bibinfo{person}{Gustav~Eje Henter}.} \bibinfo{year}{2023}\natexlab{}.
\newblock \showarticletitle{Listen, Denoise, Action! Audio-Driven Motion Synthesis with Diffusion Models}.
\newblock \bibinfo{journal}{\emph{ACM Transactions on Graphics}} \bibinfo{volume}{42}, \bibinfo{number}{4} (\bibinfo{year}{2023}).
\newblock
\showISSN{1557-7368}
\urldef\tempurl%
\url{https://doi.org/10.1145/3592458}
\showDOI{\tempurl}


\bibitem[Ao et~al\mbox{.}({[n.\,d.]})]%
        {Ao2023GestureDiffuCLIP}
\bibfield{author}{\bibinfo{person}{Tenglong Ao}, \bibinfo{person}{Zeyi Zhang}, {and} \bibinfo{person}{Libin Liu}.} \bibinfo{year}{[n.\,d.]}\natexlab{}.
\newblock \showarticletitle{GestureDiffuCLIP: Gesture Diffusion Model with CLIP Latents}.
\newblock \bibinfo{journal}{\emph{ACM Trans. Graph.}} (\bibinfo{year}{[n.\,d.]}), \bibinfo{numpages}{18}~pages.
\newblock
\urldef\tempurl%
\url{https://doi.org/10.1145/3592097}
\showDOI{\tempurl}


\bibitem[Bahdanau et~al\mbox{.}(2014)]%
        {Bahdanau2014NeuralMT}
\bibfield{author}{\bibinfo{person}{Dzmitry Bahdanau}, \bibinfo{person}{Kyunghyun Cho}, {and} \bibinfo{person}{Yoshua Bengio}.} \bibinfo{year}{2014}\natexlab{}.
\newblock \showarticletitle{Neural Machine Translation by Jointly Learning to Align and Translate}.
\newblock \bibinfo{journal}{\emph{CoRR}}  \bibinfo{volume}{abs/1409.0473} (\bibinfo{year}{2014}).
\newblock
\urldef\tempurl%
\url{https://api.semanticscholar.org/CorpusID:11212020}
\showURL{%
\tempurl}


\bibitem[Bartneck et~al\mbox{.}(2009)]%
        {Bartneck2009}
\bibfield{author}{\bibinfo{person}{Christoph Bartneck}, \bibinfo{person}{Dana Kuli{\'{c}}}, \bibinfo{person}{Elizabeth Croft}, {and} \bibinfo{person}{Susana Zoghbi}.} \bibinfo{year}{2009}\natexlab{}.
\newblock \showarticletitle{Measurement Instruments for the Anthropomorphism, Animacy, Likeability, Perceived Intelligence, and Perceived Safety of Robots}.
\newblock \bibinfo{journal}{\emph{International Journal of Social Robotics}} \bibinfo{volume}{1}, \bibinfo{number}{1} (\bibinfo{date}{01 Jan} \bibinfo{year}{2009}), \bibinfo{pages}{71--81}.
\newblock
\showISSN{1875-4805}
\urldef\tempurl%
\url{https://doi.org/10.1007/s12369-008-0001-3}
\showDOI{\tempurl}


\bibitem[Bhattacharya et~al\mbox{.}(2021)]%
        {bhattacharya2021speech}
\bibfield{author}{\bibinfo{person}{Uttaran Bhattacharya}, \bibinfo{person}{Elizabeth Childs}, \bibinfo{person}{Nicholas Rewkowski}, {and} \bibinfo{person}{Dinesh Manocha}.} \bibinfo{year}{2021}\natexlab{}.
\newblock \showarticletitle{Speech2AffectiveGestures: Synthesizing Co-Speech Gestures with Generative Adversarial Affective Expression Learning}.
\newblock \bibinfo{journal}{\emph{CoRR}}  \bibinfo{volume}{abs/2108.00262} (\bibinfo{year}{2021}).
\newblock
\showeprint[arXiv]{2108.00262}
\urldef\tempurl%
\url{https://arxiv.org/abs/2108.00262}
\showURL{%
\tempurl}


\bibitem[{Cao} et~al\mbox{.}(2019)]%
        {cao2019Openpose}
\bibfield{author}{\bibinfo{person}{Z. {Cao}}, \bibinfo{person}{G. {Hidalgo Martinez}}, \bibinfo{person}{T. {Simon}}, \bibinfo{person}{S. {Wei}}, {and} \bibinfo{person}{Y.~A. {Sheikh}}.} \bibinfo{year}{2019}\natexlab{}.
\newblock \showarticletitle{OpenPose: Realtime Multi-Person 2D Pose Estimation using Part Affinity Fields}.
\newblock \bibinfo{journal}{\emph{IEEE Transactions on Pattern Analysis and Machine Intelligence}} (\bibinfo{year}{2019}).
\newblock


\bibitem[Chemburkar et~al\mbox{.}(2023)]%
        {chemburkar2023discrete}
\bibfield{author}{\bibinfo{person}{Ankur Chemburkar}, \bibinfo{person}{Shuhong Lu}, {and} \bibinfo{person}{Andrew Feng}.} \bibinfo{year}{2023}\natexlab{}.
\newblock \showarticletitle{Discrete Diffusion for Co-Speech Gesture Synthesis}. In \bibinfo{booktitle}{\emph{Companion Publication of the 25th International Conference on Multimodal Interaction}} (, Paris, France,) \emph{(\bibinfo{series}{ICMI '23 Companion})}. \bibinfo{publisher}{Association for Computing Machinery}, \bibinfo{address}{New York, NY, USA}, \bibinfo{pages}{186–192}.
\newblock
\showISBNx{9798400703218}
\urldef\tempurl%
\url{https://doi.org/10.1145/3610661.3616556}
\showDOI{\tempurl}


\bibitem[Choutas et~al\mbox{.}(2020)]%
        {ExPose:2020}
\bibfield{author}{\bibinfo{person}{Vasileios Choutas}, \bibinfo{person}{Georgios Pavlakos}, \bibinfo{person}{Timo Bolkart}, \bibinfo{person}{Dimitrios Tzionas}, {and} \bibinfo{person}{Michael~J. Black}.} \bibinfo{year}{2020}\natexlab{}.
\newblock \showarticletitle{Monocular Expressive Body Regression through Body-Driven Attention}. In \bibinfo{booktitle}{\emph{European Conference on Computer Vision (ECCV)}}.
\newblock
\urldef\tempurl%
\url{https://expose.is.tue.mpg.de}
\showURL{%
\tempurl}


\bibitem[Clark et~al\mbox{.}(2018)]%
        {clark2018whyrate}
\bibfield{author}{\bibinfo{person}{Andrew~P. Clark}, \bibinfo{person}{Kate~L. Howard}, \bibinfo{person}{Andy~T. Woods}, \bibinfo{person}{Ian~S. Penton-Voak}, {and} \bibinfo{person}{Christof Neumann}.} \bibinfo{year}{2018}\natexlab{}.
\newblock \showarticletitle{Why rate when you could compare? Using the “EloChoice” package to assess pairwise comparisons of perceived physical strength}.
\newblock \bibinfo{journal}{\emph{PLOS ONE}} \bibinfo{volume}{13}, \bibinfo{number}{1} (\bibinfo{date}{01} \bibinfo{year}{2018}), \bibinfo{pages}{1--16}.
\newblock
\urldef\tempurl%
\url{https://doi.org/10.1371/journal.pone.0190393}
\showDOI{\tempurl}


\bibitem[Deichler et~al\mbox{.}(2023)]%
        {deichler2023diffusion}
\bibfield{author}{\bibinfo{person}{Anna Deichler}, \bibinfo{person}{Shivam Mehta}, \bibinfo{person}{Simon Alexanderson}, {and} \bibinfo{person}{Jonas Beskow}.} \bibinfo{year}{2023}\natexlab{}.
\newblock \showarticletitle{Diffusion-Based Co-Speech Gesture Generation Using Joint Text and Audio Representation}. In \bibinfo{booktitle}{\emph{Proceedings of the 25th International Conference on Multimodal Interaction}} \emph{(\bibinfo{series}{ICMI '23})}. \bibinfo{publisher}{Association for Computing Machinery}, \bibinfo{address}{New York, NY, USA}, \bibinfo{pages}{755–762}.
\newblock
\showISBNx{9798400700552}
\urldef\tempurl%
\url{https://doi.org/10.1145/3577190.3616117}
\showDOI{\tempurl}


\bibitem[Douglas et~al\mbox{.}(2023)]%
        {Douglas2023-vo}
\bibfield{author}{\bibinfo{person}{Benjamin~D Douglas}, \bibinfo{person}{Patrick~J Ewell}, {and} \bibinfo{person}{Markus Brauer}.} \bibinfo{year}{2023}\natexlab{}.
\newblock \showarticletitle{Data quality in online human-subjects research: Comparisons between {MTurk}, Prolific, {CloudResearch}, Qualtrics, and {SONA}}.
\newblock \bibinfo{journal}{\emph{PLoS One}} \bibinfo{volume}{18}, \bibinfo{number}{3} (\bibinfo{date}{March} \bibinfo{year}{2023}), \bibinfo{pages}{e0279720}.
\newblock


\bibitem[Fares et~al\mbox{.}(2023)]%
        {fares2023zero}
\bibfield{author}{\bibinfo{person}{Mireille Fares}, \bibinfo{person}{Catherine Pelachaud}, {and} \bibinfo{person}{Nicolas Obin}.} \bibinfo{year}{2023}\natexlab{}.
\newblock \showarticletitle{Zero-Shot Style Transfer for Multimodal Data-Driven Gesture Synthesis}. In \bibinfo{booktitle}{\emph{2023 IEEE 17th International Conference on Automatic Face and Gesture Recognition (FG)}}. \bibinfo{pages}{1--4}.
\newblock
\urldef\tempurl%
\url{https://doi.org/10.1109/FG57933.2023.10042658}
\showDOI{\tempurl}


\bibitem[Ferstl and McDonnell(2018)]%
        {ferstl2018trinity}
\bibfield{author}{\bibinfo{person}{Ylva Ferstl} {and} \bibinfo{person}{Rachel McDonnell}.} \bibinfo{year}{2018}\natexlab{}.
\newblock \showarticletitle{Investigating the use of recurrent motion modelling for speech gesture generation}. In \bibinfo{booktitle}{\emph{Proceedings of the 18th International Conference on Intelligent Virtual Agents}} (Sydney, NSW, Australia) \emph{(\bibinfo{series}{IVA '18})}. \bibinfo{publisher}{Association for Computing Machinery}, \bibinfo{address}{New York, NY, USA}, \bibinfo{pages}{93–98}.
\newblock
\showISBNx{9781450360135}
\urldef\tempurl%
\url{https://doi.org/10.1145/3267851.3267898}
\showDOI{\tempurl}


\bibitem[Ferstl et~al\mbox{.}(2021)]%
        {frestl2021trinity2}
\bibfield{author}{\bibinfo{person}{Ylva Ferstl}, \bibinfo{person}{Michael Neff}, {and} \bibinfo{person}{Rachel McDonnell}.} \bibinfo{year}{2021}\natexlab{}.
\newblock \showarticletitle{ExpressGesture: Expressive gesture generation from speech through database matching}.
\newblock \bibinfo{journal}{\emph{Computer Animation and Virtual Worlds}} \bibinfo{volume}{32}, \bibinfo{number}{3-4} (\bibinfo{year}{2021}), \bibinfo{pages}{e2016}.
\newblock
\urldef\tempurl%
\url{https://doi.org/10.1002/cav.2016}
\showDOI{\tempurl}
\showeprint{https://onlinelibrary.wiley.com/doi/pdf/10.1002/cav.2016}


\bibitem[Garrido et~al\mbox{.}(2016)]%
        {garrido2016reconstruction}
\bibfield{author}{\bibinfo{person}{Pablo Garrido}, \bibinfo{person}{Michael Zollh\"{o}fer}, \bibinfo{person}{Dan Casas}, \bibinfo{person}{Levi Valgaerts}, \bibinfo{person}{Kiran Varanasi}, \bibinfo{person}{Patrick P\'{e}rez}, {and} \bibinfo{person}{Christian Theobalt}.} \bibinfo{year}{2016}\natexlab{}.
\newblock \showarticletitle{Reconstruction of Personalized 3D Face Rigs from Monocular Video}.
\newblock \bibinfo{journal}{\emph{ACM Trans. Graph.}} \bibinfo{volume}{35}, \bibinfo{number}{3}, Article \bibinfo{articleno}{28} (\bibinfo{date}{may} \bibinfo{year}{2016}), \bibinfo{numpages}{15}~pages.
\newblock
\showISSN{0730-0301}
\urldef\tempurl%
\url{https://doi.org/10.1145/2890493}
\showDOI{\tempurl}


\bibitem[Ghorbani et~al\mbox{.}(2022)]%
        {ghorbani2022zeroeggs}
\bibfield{author}{\bibinfo{person}{Saeed Ghorbani}, \bibinfo{person}{Ylva Ferstl}, \bibinfo{person}{Daniel Holden}, \bibinfo{person}{Nikolaus~F. Troje}, {and} \bibinfo{person}{Marc-André Carbonneau}.} \bibinfo{year}{2022}\natexlab{}.
\newblock \bibinfo{title}{ZeroEGGS: Zero-shot Example-based Gesture Generation from Speech}.
\newblock
\newblock
\showeprint[arxiv]{2209.07556}~[cs.GR]


\bibitem[Ginosar et~al\mbox{.}(2019)]%
        {ginosar2019learning}
\bibfield{author}{\bibinfo{person}{Shiry Ginosar}, \bibinfo{person}{Amir Bar}, \bibinfo{person}{Gefen Kohavi}, \bibinfo{person}{Caroline Chan}, \bibinfo{person}{Andrew Owens}, {and} \bibinfo{person}{Jitendra Malik}.} \bibinfo{year}{2019}\natexlab{}.
\newblock \showarticletitle{Learning Individual Styles of Conversational Gesture}.
\newblock \bibinfo{journal}{\emph{CoRR}}  \bibinfo{volume}{abs/1906.04160} (\bibinfo{year}{2019}).
\newblock
\showeprint[arXiv]{1906.04160}
\urldef\tempurl%
\url{http://arxiv.org/abs/1906.04160}
\showURL{%
\tempurl}


\bibitem[Gulati et~al\mbox{.}(2020)]%
        {gulati2020conformer}
\bibfield{author}{\bibinfo{person}{Anmol Gulati}, \bibinfo{person}{James Qin}, \bibinfo{person}{Chung-Cheng Chiu}, \bibinfo{person}{Niki Parmar}, \bibinfo{person}{Yu Zhang}, \bibinfo{person}{Jiahui Yu}, \bibinfo{person}{Wei Han}, \bibinfo{person}{Shibo Wang}, \bibinfo{person}{Zhengdong Zhang}, \bibinfo{person}{Yonghui Wu}, {and} \bibinfo{person}{Ruoming Pang}.} \bibinfo{year}{2020}\natexlab{}.
\newblock \bibinfo{title}{Conformer: Convolution-augmented Transformer for Speech Recognition}.
\newblock
\newblock
\showeprint[arxiv]{2005.08100}~[eess.AS]


\bibitem[Habibie et~al\mbox{.}(2021)]%
        {habibie2021learning}
\bibfield{author}{\bibinfo{person}{Ikhsanul Habibie}, \bibinfo{person}{Weipeng Xu}, \bibinfo{person}{Dushyant Mehta}, \bibinfo{person}{Lingjie Liu}, \bibinfo{person}{Hans-Peter Seidel}, \bibinfo{person}{Gerard Pons-Moll}, \bibinfo{person}{Mohamed Elgharib}, {and} \bibinfo{person}{Christian Theobalt}.} \bibinfo{year}{2021}\natexlab{}.
\newblock \showarticletitle{Learning Speech-driven 3D Conversational Gestures from Video}. In \bibinfo{booktitle}{\emph{ACM International Conference on Intelligent Virtual Agents (IVA)}}.
\newblock
\showeprint{Todo}


\bibitem[Heusel et~al\mbox{.}(2017)]%
        {gans2017Heusel}
\bibfield{author}{\bibinfo{person}{Martin Heusel}, \bibinfo{person}{Hubert Ramsauer}, \bibinfo{person}{Thomas Unterthiner}, \bibinfo{person}{Bernhard Nessler}, \bibinfo{person}{G{\"{u}}nter Klambauer}, {and} \bibinfo{person}{Sepp Hochreiter}.} \bibinfo{year}{2017}\natexlab{}.
\newblock \showarticletitle{GANs Trained by a Two Time-Scale Update Rule Converge to a Nash Equilibrium}.
\newblock \bibinfo{journal}{\emph{CoRR}}  \bibinfo{volume}{abs/1706.08500} (\bibinfo{year}{2017}).
\newblock
\showeprint[arXiv]{1706.08500}
\urldef\tempurl%
\url{http://arxiv.org/abs/1706.08500}
\showURL{%
\tempurl}


\bibitem[Ho et~al\mbox{.}(2020)]%
        {ho2020diffusion}
\bibfield{author}{\bibinfo{person}{Jonathan Ho}, \bibinfo{person}{Ajay Jain}, {and} \bibinfo{person}{Pieter Abbeel}.} \bibinfo{year}{2020}\natexlab{}.
\newblock \showarticletitle{Denoising diffusion probabilistic models}. In \bibinfo{booktitle}{\emph{Proceedings of the 34th International Conference on Neural Information Processing Systems}} (Vancouver, BC, Canada) \emph{(\bibinfo{series}{NIPS'20})}. \bibinfo{publisher}{Curran Associates Inc.}, \bibinfo{address}{Red Hook, NY, USA}, Article \bibinfo{articleno}{574}, \bibinfo{numpages}{12}~pages.
\newblock
\showISBNx{9781713829546}


\bibitem[Ho and Salimans(2022)]%
        {ho2022classifierfree}
\bibfield{author}{\bibinfo{person}{Jonathan Ho} {and} \bibinfo{person}{Tim Salimans}.} \bibinfo{year}{2022}\natexlab{}.
\newblock \bibinfo{title}{Classifier-Free Diffusion Guidance}.
\newblock
\newblock
\showeprint[arxiv]{2207.12598}~[cs.LG]


\bibitem[Huang and Belongie(2017)]%
        {huang2017adain}
\bibfield{author}{\bibinfo{person}{Xun Huang} {and} \bibinfo{person}{Serge~J. Belongie}.} \bibinfo{year}{2017}\natexlab{}.
\newblock \showarticletitle{Arbitrary Style Transfer in Real-time with Adaptive Instance Normalization}.
\newblock \bibinfo{journal}{\emph{CoRR}}  \bibinfo{volume}{abs/1703.06868} (\bibinfo{year}{2017}).
\newblock
\showeprint[arXiv]{1703.06868}
\urldef\tempurl%
\url{http://arxiv.org/abs/1703.06868}
\showURL{%
\tempurl}


\bibitem[Ionescu et~al\mbox{.}(2014)]%
        {h36m_pami}
\bibfield{author}{\bibinfo{person}{Catalin Ionescu}, \bibinfo{person}{Dragos Papava}, \bibinfo{person}{Vlad Olaru}, {and} \bibinfo{person}{Cristian Sminchisescu}.} \bibinfo{year}{2014}\natexlab{}.
\newblock \showarticletitle{Human3.6M: Large Scale Datasets and Predictive Methods for 3D Human Sensing in Natural Environments}.
\newblock \bibinfo{journal}{\emph{IEEE Transactions on Pattern Analysis and Machine Intelligence}} \bibinfo{volume}{36}, \bibinfo{number}{7} (\bibinfo{date}{jul} \bibinfo{year}{2014}), \bibinfo{pages}{1325--1339}.
\newblock


\bibitem[Jonell et~al\mbox{.}(2020)]%
        {Jonell_2020}
\bibfield{author}{\bibinfo{person}{Patrik Jonell}, \bibinfo{person}{Taras Kucherenko}, \bibinfo{person}{Ilaria Torre}, {and} \bibinfo{person}{Jonas Beskow}.} \bibinfo{year}{2020}\natexlab{}.
\newblock \showarticletitle{Can we trust online crowdworkers?: Comparing online and offline participants in a preference test of virtual agents}. In \bibinfo{booktitle}{\emph{Proceedings of the 20th ACM International Conference on Intelligent Virtual Agents}} \emph{(\bibinfo{series}{IVA ’20})}. \bibinfo{publisher}{ACM}.
\newblock
\urldef\tempurl%
\url{https://doi.org/10.1145/3383652.3423860}
\showDOI{\tempurl}


\bibitem[Kucherenko et~al\mbox{.}(2020a)]%
        {kucherenko2020gesticulator}
\bibfield{author}{\bibinfo{person}{Taras Kucherenko}, \bibinfo{person}{Patrik Jonell}, \bibinfo{person}{Sanne van Waveren}, \bibinfo{person}{Gustav~Eje Henter}, \bibinfo{person}{Simon Alexanderson}, \bibinfo{person}{Iolanda Leite}, {and} \bibinfo{person}{Hedvig Kjellstr{\"{o}}m}.} \bibinfo{year}{2020}\natexlab{a}.
\newblock \showarticletitle{Gesticulator: {A} framework for semantically-aware speech-driven gesture generation}.
\newblock \bibinfo{journal}{\emph{CoRR}}  \bibinfo{volume}{abs/2001.09326} (\bibinfo{year}{2020}).
\newblock
\showeprint[arXiv]{2001.09326}
\urldef\tempurl%
\url{https://arxiv.org/abs/2001.09326}
\showURL{%
\tempurl}


\bibitem[Kucherenko et~al\mbox{.}(2020b)]%
        {kucherenko2020genea}
\bibfield{author}{\bibinfo{person}{Taras Kucherenko}, \bibinfo{person}{Patrik Jonell}, \bibinfo{person}{Youngwoo Yoon}, \bibinfo{person}{Pieter Wolfert}, {and} \bibinfo{person}{Gustav~Eje Henter}.} \bibinfo{year}{2020}\natexlab{b}.
\newblock \bibinfo{title}{{The GENEA Challenge 2020: Benchmarking gesture- generation systems on common data}}.
\newblock
\newblock
\urldef\tempurl%
\url{https://doi.org/10.5281/zenodo.4094697}
\showDOI{\tempurl}


\bibitem[Kucherenko et~al\mbox{.}(2023)]%
        {kucherenko2023genea}
\bibfield{author}{\bibinfo{person}{Taras Kucherenko}, \bibinfo{person}{Rajmund Nagy}, \bibinfo{person}{Youngwoo Yoon}, \bibinfo{person}{Jieyeon Woo}, \bibinfo{person}{Teodor Nikolov}, \bibinfo{person}{Mihail Tsakov}, {and} \bibinfo{person}{Gustav~Eje Henter}.} \bibinfo{year}{2023}\natexlab{}.
\newblock \showarticletitle{The {GENEA} {C}hallenge 2023: {A} large-scale evaluation of gesture generation models in monadic and dyadic settings}. In \bibinfo{booktitle}{\emph{Proceedings of the ACM International Conference on Multimodal Interaction}} \emph{(\bibinfo{series}{ICMI '23})}. \bibinfo{publisher}{ACM}.
\newblock


\bibitem[Lee et~al\mbox{.}(2019a)]%
        {lee2019talking}
\bibfield{author}{\bibinfo{person}{Gilwoo Lee}, \bibinfo{person}{Zhiwei Deng}, \bibinfo{person}{Shugao Ma}, \bibinfo{person}{Takaaki Shiratori}, \bibinfo{person}{Siddhartha Srinivasa}, {and} \bibinfo{person}{Yaser Sheikh}.} \bibinfo{year}{2019}\natexlab{a}.
\newblock \showarticletitle{Talking With Hands 16.2M: A Large-Scale Dataset of Synchronized Body-Finger Motion and Audio for Conversational Motion Analysis and Synthesis}. In \bibinfo{booktitle}{\emph{2019 IEEE/CVF International Conference on Computer Vision (ICCV)}}. \bibinfo{pages}{763--772}.
\newblock
\urldef\tempurl%
\url{https://doi.org/10.1109/ICCV.2019.00085}
\showDOI{\tempurl}


\bibitem[Lee et~al\mbox{.}(2019b)]%
        {hsin2019diversity}
\bibfield{author}{\bibinfo{person}{Hsin{-}Ying Lee}, \bibinfo{person}{Xiaodong Yang}, \bibinfo{person}{Ming{-}Yu Liu}, \bibinfo{person}{Ting{-}Chun Wang}, \bibinfo{person}{Yu{-}Ding Lu}, \bibinfo{person}{Ming{-}Hsuan Yang}, {and} \bibinfo{person}{Jan Kautz}.} \bibinfo{year}{2019}\natexlab{b}.
\newblock \showarticletitle{Dancing to Music}.
\newblock \bibinfo{journal}{\emph{CoRR}}  \bibinfo{volume}{abs/1911.02001} (\bibinfo{year}{2019}).
\newblock
\showeprint[arXiv]{1911.02001}
\urldef\tempurl%
\url{http://arxiv.org/abs/1911.02001}
\showURL{%
\tempurl}


\bibitem[Li et~al\mbox{.}(2021c)]%
        {dance2021Li}
\bibfield{author}{\bibinfo{person}{Buyu Li}, \bibinfo{person}{Yongchi Zhao}, {and} \bibinfo{person}{Lu Sheng}.} \bibinfo{year}{2021}\natexlab{c}.
\newblock \showarticletitle{DanceNet3D: Music Based Dance Generation with Parametric Motion Transformer}.
\newblock \bibinfo{journal}{\emph{CoRR}}  \bibinfo{volume}{abs/2103.10206} (\bibinfo{year}{2021}).
\newblock
\showeprint[arXiv]{2103.10206}
\urldef\tempurl%
\url{https://arxiv.org/abs/2103.10206}
\showURL{%
\tempurl}


\bibitem[Li et~al\mbox{.}(2021a)]%
        {li2021audio}
\bibfield{author}{\bibinfo{person}{Jing Li}, \bibinfo{person}{Di Kang}, \bibinfo{person}{Wenjie Pei}, \bibinfo{person}{Xuefei Zhe}, \bibinfo{person}{Ying Zhang}, \bibinfo{person}{Zhenyu He}, {and} \bibinfo{person}{Linchao Bao}.} \bibinfo{year}{2021}\natexlab{a}.
\newblock \showarticletitle{Audio2Gestures: Generating Diverse Gestures from Speech Audio with Conditional Variational Autoencoders}.
\newblock \bibinfo{journal}{\emph{CoRR}}  \bibinfo{volume}{abs/2108.06720} (\bibinfo{year}{2021}).
\newblock
\showeprint[arXiv]{2108.06720}
\urldef\tempurl%
\url{https://arxiv.org/abs/2108.06720}
\showURL{%
\tempurl}


\bibitem[Li et~al\mbox{.}(2021b)]%
        {learn2021Li}
\bibfield{author}{\bibinfo{person}{Ruilong Li}, \bibinfo{person}{Shan Yang}, \bibinfo{person}{David~A. Ross}, {and} \bibinfo{person}{Angjoo Kanazawa}.} \bibinfo{year}{2021}\natexlab{b}.
\newblock \showarticletitle{Learn to Dance with {AIST++:} Music Conditioned 3D Dance Generation}.
\newblock \bibinfo{journal}{\emph{CoRR}}  \bibinfo{volume}{abs/2101.08779} (\bibinfo{year}{2021}).
\newblock
\showeprint[arXiv]{2101.08779}
\urldef\tempurl%
\url{https://arxiv.org/abs/2101.08779}
\showURL{%
\tempurl}


\bibitem[Liu et~al\mbox{.}(2022b)]%
        {liu2022beat}
\bibfield{author}{\bibinfo{person}{Haiyang Liu}, \bibinfo{person}{Zihao Zhu}, \bibinfo{person}{Naoya Iwamoto}, \bibinfo{person}{Yichen Peng}, \bibinfo{person}{Zhengqing Li}, \bibinfo{person}{You Zhou}, \bibinfo{person}{Elif Bozkurt}, {and} \bibinfo{person}{Bo Zheng}.} \bibinfo{year}{2022}\natexlab{b}.
\newblock \bibinfo{title}{BEAT: A Large-Scale Semantic and Emotional Multi-Modal Dataset for Conversational Gestures Synthesis}.
\newblock
\newblock
\showeprint[arxiv]{2203.05297}~[cs.CV]


\bibitem[Liu et~al\mbox{.}(2022a)]%
        {liu2022learning}
\bibfield{author}{\bibinfo{person}{Xian Liu}, \bibinfo{person}{Qianyi Wu}, \bibinfo{person}{Hang Zhou}, \bibinfo{person}{Yinghao Xu}, \bibinfo{person}{Rui Qian}, \bibinfo{person}{Xinyi Lin}, \bibinfo{person}{Xiaowei Zhou}, \bibinfo{person}{Wayne Wu}, \bibinfo{person}{Bo Dai}, {and} \bibinfo{person}{Bolei Zhou}.} \bibinfo{year}{2022}\natexlab{a}.
\newblock \showarticletitle{Learning Hierarchical Cross-Modal Association for Co-Speech Gesture Generation}. In \bibinfo{booktitle}{\emph{Proceedings of the IEEE/CVF Conference on Computer Vision and Pattern Recognition}}. \bibinfo{pages}{10462--10472}.
\newblock


\bibitem[Mcneill(1992)]%
        {mcneill1994hand}
\bibfield{author}{\bibinfo{person}{David Mcneill}.} \bibinfo{year}{1992}\natexlab{}.
\newblock \showarticletitle{Hand and Mind: What Gestures Reveal About Thought}.
\newblock \bibinfo{journal}{\emph{University of Chicago Press}}  \bibinfo{volume}{27} (\bibinfo{year}{1992}).
\newblock
\urldef\tempurl%
\url{https://doi.org/10.2307/1576015}
\showDOI{\tempurl}


\bibitem[Mehta et~al\mbox{.}(2019)]%
        {mehta2019XNect}
\bibfield{author}{\bibinfo{person}{Dushyant Mehta}, \bibinfo{person}{Oleksandr Sotnychenko}, \bibinfo{person}{Franziska Mueller}, \bibinfo{person}{Weipeng Xu}, \bibinfo{person}{Mohamed Elgharib}, \bibinfo{person}{Pascal Fua}, \bibinfo{person}{Hans{-}Peter Seidel}, \bibinfo{person}{Helge Rhodin}, \bibinfo{person}{Gerard Pons{-}Moll}, {and} \bibinfo{person}{Christian Theobalt}.} \bibinfo{year}{2019}\natexlab{}.
\newblock \showarticletitle{XNect: Real-time Multi-person 3D Human Pose Estimation with a Single {RGB} Camera}.
\newblock \bibinfo{journal}{\emph{CoRR}}  \bibinfo{volume}{abs/1907.00837} (\bibinfo{year}{2019}).
\newblock
\showeprint[arXiv]{1907.00837}
\urldef\tempurl%
\url{http://arxiv.org/abs/1907.00837}
\showURL{%
\tempurl}


\bibitem[Mehta et~al\mbox{.}(2023)]%
        {Mehta_2023}
\bibfield{author}{\bibinfo{person}{Shivam Mehta}, \bibinfo{person}{Siyang Wang}, \bibinfo{person}{Simon Alexanderson}, \bibinfo{person}{Jonas Beskow}, \bibinfo{person}{Eva Szekely}, {and} \bibinfo{person}{Gustav~Eje Henter}.} \bibinfo{year}{2023}\natexlab{}.
\newblock \showarticletitle{Diff-TTSG: Denoising probabilistic integrated speech and gesture synthesis}. In \bibinfo{booktitle}{\emph{12th ISCA Speech Synthesis Workshop (SSW2023)}}. \bibinfo{publisher}{ISCA}.
\newblock
\urldef\tempurl%
\url{https://doi.org/10.21437/ssw.2023-24}
\showDOI{\tempurl}


\bibitem[Nyatsanga et~al\mbox{.}(2023)]%
        {Nyatsanga_2023}
\bibfield{author}{\bibinfo{person}{S. Nyatsanga}, \bibinfo{person}{T. Kucherenko}, \bibinfo{person}{C. Ahuja}, \bibinfo{person}{G.~E. Henter}, {and} \bibinfo{person}{M. Neff}.} \bibinfo{year}{2023}\natexlab{}.
\newblock \showarticletitle{A Comprehensive Review of Data‐Driven Co‐Speech Gesture Generation}.
\newblock \bibinfo{journal}{\emph{Computer Graphics Forum}} \bibinfo{volume}{42}, \bibinfo{number}{2} (\bibinfo{date}{May} \bibinfo{year}{2023}), \bibinfo{pages}{569–596}.
\newblock
\showISSN{1467-8659}
\urldef\tempurl%
\url{https://doi.org/10.1111/cgf.14776}
\showDOI{\tempurl}


\bibitem[Pavllo et~al\mbox{.}(2019)]%
        {pavllo:videopose3d:2019}
\bibfield{author}{\bibinfo{person}{Dario Pavllo}, \bibinfo{person}{Christoph Feichtenhofer}, \bibinfo{person}{David Grangier}, {and} \bibinfo{person}{Michael Auli}.} \bibinfo{year}{2019}\natexlab{}.
\newblock \showarticletitle{3D human pose estimation in video with temporal convolutions and semi-supervised training}. In \bibinfo{booktitle}{\emph{Conference on Computer Vision and Pattern Recognition (CVPR)}}.
\newblock


\bibitem[Qian et~al\mbox{.}(2021)]%
        {qian2021speech}
\bibfield{author}{\bibinfo{person}{Shenhan Qian}, \bibinfo{person}{Zhi Tu}, \bibinfo{person}{Yihao Zhi}, \bibinfo{person}{Wen Liu}, {and} \bibinfo{person}{Shenghua Gao}.} \bibinfo{year}{2021}\natexlab{}.
\newblock \showarticletitle{Speech Drives Templates: Co-Speech Gesture Synthesis with Learned Templates}. In \bibinfo{booktitle}{\emph{2021 IEEE/CVF International Conference on Computer Vision (ICCV)}}. IEEE, \bibinfo{pages}{11057--11066}.
\newblock


\bibitem[Radford et~al\mbox{.}(2021)]%
        {clip2021radford}
\bibfield{author}{\bibinfo{person}{Alec Radford}, \bibinfo{person}{Jong~Wook Kim}, \bibinfo{person}{Chris Hallacy}, \bibinfo{person}{Aditya Ramesh}, \bibinfo{person}{Gabriel Goh}, \bibinfo{person}{Sandhini Agarwal}, \bibinfo{person}{Girish Sastry}, \bibinfo{person}{Amanda Askell}, \bibinfo{person}{Pamela Mishkin}, \bibinfo{person}{Jack Clark}, \bibinfo{person}{Gretchen Krueger}, {and} \bibinfo{person}{Ilya Sutskever}.} \bibinfo{year}{2021}\natexlab{}.
\newblock \showarticletitle{Learning Transferable Visual Models From Natural Language Supervision}.
\newblock \bibinfo{journal}{\emph{CoRR}}  \bibinfo{volume}{abs/2103.00020} (\bibinfo{year}{2021}).
\newblock
\showeprint[arXiv]{2103.00020}
\urldef\tempurl%
\url{https://arxiv.org/abs/2103.00020}
\showURL{%
\tempurl}


\bibitem[Sohl{-}Dickstein et~al\mbox{.}(2015)]%
        {sohl2015deep}
\bibfield{author}{\bibinfo{person}{Jascha Sohl{-}Dickstein}, \bibinfo{person}{Eric~A. Weiss}, \bibinfo{person}{Niru Maheswaranathan}, {and} \bibinfo{person}{Surya Ganguli}.} \bibinfo{year}{2015}\natexlab{}.
\newblock \showarticletitle{Deep Unsupervised Learning using Nonequilibrium Thermodynamics}.
\newblock \bibinfo{journal}{\emph{CoRR}}  \bibinfo{volume}{abs/1503.03585} (\bibinfo{year}{2015}).
\newblock
\showeprint[arXiv]{1503.03585}
\urldef\tempurl%
\url{http://arxiv.org/abs/1503.03585}
\showURL{%
\tempurl}


\bibitem[Song and Ermon(2019)]%
        {song2019generative}
\bibfield{author}{\bibinfo{person}{Yang Song} {and} \bibinfo{person}{Stefano Ermon}.} \bibinfo{year}{2019}\natexlab{}.
\newblock \showarticletitle{Generative Modeling by Estimating Gradients of the Data Distribution}.
\newblock \bibinfo{journal}{\emph{CoRR}}  \bibinfo{volume}{abs/1907.05600} (\bibinfo{year}{2019}).
\newblock
\showeprint[arXiv]{1907.05600}
\urldef\tempurl%
\url{http://arxiv.org/abs/1907.05600}
\showURL{%
\tempurl}


\bibitem[Tonoli et~al\mbox{.}(2023)]%
        {tonoli2023gesture}
\bibfield{author}{\bibinfo{person}{Rodolfo~Luis Tonoli}, \bibinfo{person}{Leonardo~Boulitreau de Menezes Martins~Marques}, \bibinfo{person}{Lucas~Hideki Ueda}, {and} \bibinfo{person}{Paula Paro~Dornhofer Costa}.} \bibinfo{year}{2023}\natexlab{}.
\newblock \showarticletitle{Gesture Generation with Diffusion Models Aided by Speech Activity Information}. In \bibinfo{booktitle}{\emph{GENEA: Generation and Evaluation of Non-verbal Behaviour for Embodied Agents Challenge 2023}}.
\newblock
\urldef\tempurl%
\url{https://openreview.net/forum?id=S9Efb3MoiZ}
\showURL{%
\tempurl}


\bibitem[Vo{\ss} and Kopp(2023)]%
        {voss2023aq}
\bibfield{author}{\bibinfo{person}{Hendric Vo{\ss}} {and} \bibinfo{person}{Stefan Kopp}.} \bibinfo{year}{2023}\natexlab{}.
\newblock \showarticletitle{AQ-GT: a Temporally Aligned and Quantized GRU-Transformer for Co-Speech Gesture Synthesis}.
\newblock \bibinfo{journal}{\emph{arXiv preprint arXiv:2305.01241}} (\bibinfo{year}{2023}).
\newblock


\bibitem[Windle et~al\mbox{.}(2023)]%
        {windle2023the}
\bibfield{author}{\bibinfo{person}{Jonathan Windle}, \bibinfo{person}{Iain Matthews}, \bibinfo{person}{Ben Milner}, {and} \bibinfo{person}{Sarah Taylor}.} \bibinfo{year}{2023}\natexlab{}.
\newblock \showarticletitle{The {UEA} Digital Humans entry to the {GENEA} Challenge 2023}. In \bibinfo{booktitle}{\emph{GENEA: Generation and Evaluation of Non-verbal Behaviour for Embodied Agents Challenge 2023}}.
\newblock
\urldef\tempurl%
\url{https://openreview.net/forum?id=bBrebR1YpXe}
\showURL{%
\tempurl}


\bibitem[Wolfert et~al\mbox{.}(2022)]%
        {Wolfert_2022}
\bibfield{author}{\bibinfo{person}{Pieter Wolfert}, \bibinfo{person}{Nicole Robinson}, {and} \bibinfo{person}{Tony Belpaeme}.} \bibinfo{year}{2022}\natexlab{}.
\newblock \showarticletitle{A Review of Evaluation Practices of Gesture Generation in Embodied Conversational Agents}.
\newblock \bibinfo{journal}{\emph{IEEE Transactions on Human-Machine Systems}} \bibinfo{volume}{52}, \bibinfo{number}{3} (\bibinfo{date}{June} \bibinfo{year}{2022}), \bibinfo{pages}{379–389}.
\newblock
\showISSN{2168-2305}
\urldef\tempurl%
\url{https://doi.org/10.1109/thms.2022.3149173}
\showDOI{\tempurl}


\bibitem[Yang et~al\mbox{.}(2023b)]%
        {yang2023diffusestylegesture}
\bibfield{author}{\bibinfo{person}{Sicheng Yang}, \bibinfo{person}{Zhiyong Wu}, \bibinfo{person}{Minglei Li}, \bibinfo{person}{Zhensong Zhang}, \bibinfo{person}{Lei Hao}, \bibinfo{person}{Weihong Bao}, \bibinfo{person}{Ming Cheng}, {and} \bibinfo{person}{Long Xiao}.} \bibinfo{year}{2023}\natexlab{b}.
\newblock \bibinfo{title}{DiffuseStyleGesture: Stylized Audio-Driven Co-Speech Gesture Generation with Diffusion Models}.
\newblock
\newblock
\showeprint[arxiv]{2305.04919}~[cs.HC]


\bibitem[Yang et~al\mbox{.}(2023a)]%
        {yang2023qpgesture}
\bibfield{author}{\bibinfo{person}{Sicheng Yang}, \bibinfo{person}{Zhiyong Wu}, \bibinfo{person}{Minglei Li}, \bibinfo{person}{Zhensong Zhang}, \bibinfo{person}{Lei Hao}, \bibinfo{person}{Weihong Bao}, {and} \bibinfo{person}{Haolin Zhuang}.} \bibinfo{year}{2023}\natexlab{a}.
\newblock \bibinfo{title}{QPGesture: Quantization-Based and Phase-Guided Motion Matching for Natural Speech-Driven Gesture Generation}.
\newblock
\newblock
\showeprint[arxiv]{2305.11094}~[cs.HC]


\bibitem[Yoon et~al\mbox{.}(2020)]%
        {Yoon2020Speech}
\bibfield{author}{\bibinfo{person}{Youngwoo Yoon}, \bibinfo{person}{Bok Cha}, \bibinfo{person}{Joo-Haeng Lee}, \bibinfo{person}{Minsu Jang}, \bibinfo{person}{Jaeyeon Lee}, \bibinfo{person}{Jaehong Kim}, {and} \bibinfo{person}{Geehyuk Lee}.} \bibinfo{year}{2020}\natexlab{}.
\newblock \showarticletitle{Speech Gesture Generation from the Trimodal Context of Text, Audio, and Speaker Identity}.
\newblock \bibinfo{journal}{\emph{ACM Transactions on Graphics}} \bibinfo{volume}{39}, \bibinfo{number}{6} (\bibinfo{year}{2020}).
\newblock


\bibitem[Yoon et~al\mbox{.}(2018)]%
        {yoon2018robot}
\bibfield{author}{\bibinfo{person}{Youngwoo Yoon}, \bibinfo{person}{Woo{-}Ri Ko}, \bibinfo{person}{Minsu Jang}, \bibinfo{person}{Jaeyeon Lee}, \bibinfo{person}{Jaehong Kim}, {and} \bibinfo{person}{Geehyuk Lee}.} \bibinfo{year}{2018}\natexlab{}.
\newblock \showarticletitle{Robots Learn Social Skills: End-to-End Learning of Co-Speech Gesture Generation for Humanoid Robots}.
\newblock \bibinfo{journal}{\emph{CoRR}}  \bibinfo{volume}{abs/1810.12541} (\bibinfo{year}{2018}).
\newblock
\showeprint[arXiv]{1810.12541}
\urldef\tempurl%
\url{http://arxiv.org/abs/1810.12541}
\showURL{%
\tempurl}


\bibitem[Zhao et~al\mbox{.}(2023)]%
        {zhao2023diffugesture}
\bibfield{author}{\bibinfo{person}{Weiyu Zhao}, \bibinfo{person}{Liangxiao Hu}, {and} \bibinfo{person}{Shengping Zhang}.} \bibinfo{year}{2023}\natexlab{}.
\newblock \showarticletitle{DiffuGesture: Generating Human Gesture From Two-person Dialogue With Diffusion Models}. In \bibinfo{booktitle}{\emph{Companion Publication of the 25th International Conference on Multimodal Interaction}} (<conf-loc>, <city>Paris</city>, <country>France</country>, </conf-loc>) \emph{(\bibinfo{series}{ICMI '23 Companion})}. \bibinfo{publisher}{Association for Computing Machinery}, \bibinfo{address}{New York, NY, USA}, \bibinfo{pages}{179–185}.
\newblock
\showISBNx{9798400703218}
\urldef\tempurl%
\url{https://doi.org/10.1145/3610661.3616552}
\showDOI{\tempurl}


\bibitem[Zhu et~al\mbox{.}(2023)]%
        {zhu2023taming}
\bibfield{author}{\bibinfo{person}{Lingting Zhu}, \bibinfo{person}{Xian Liu}, \bibinfo{person}{Xuanyu Liu}, \bibinfo{person}{Rui Qian}, \bibinfo{person}{Ziwei Liu}, {and} \bibinfo{person}{Lequan Yu}.} \bibinfo{year}{2023}\natexlab{}.
\newblock \showarticletitle{Taming Diffusion Models for Audio-Driven Co-Speech Gesture Generation}. In \bibinfo{booktitle}{\emph{Proceedings of the IEEE/CVF Conference on Computer Vision and Pattern Recognition}}. \bibinfo{pages}{10544--10553}.
\newblock


\end{thebibliography}

\appendix

\end{document}